\documentclass[letterpaper]{article} 
\usepackage{aaai2027}  
\usepackage[hyphens]{url}  
\usepackage{graphicx} 
\usepackage{natbib}  
\usepackage{bibunits}
\usepackage{caption} 
\usepackage{algorithm}
\usepackage[noend]{algpseudocode}

\usepackage{newfloat}
\usepackage{listings}
\DeclareCaptionStyle{ruled}{labelfont=normalfont,labelsep=colon,strut=off} 
\floatstyle{ruled}
\newfloat{listing}{tb}{lst}{}
\floatname{listing}{Listing}

\usepackage{booktabs}
\usepackage{bm}
\usepackage{amsmath}
\usepackage{amssymb}
\usepackage{subcaption}
\usepackage{multirow}
\newcommand{\tablebodyfont}{\footnotesize}

\nocopyright

\title{Inference-Time Trajectory Optimization \\for Structure-Preserving Manga Image Editing}
\author{
    Ryosuke Furuta
}
\affiliations{
    Mantra Inc.\\
    furuta@mantra.co.jp
}

\begin{document}

\makeatletter
\let\std@cite\citep
\makeatother
\maketitle
\begin{bibunit}[aaai2027]

\begin{abstract}
We present a lightweight, training-free trajectory correction method that adapts a pretrained image editing model to each input manga image using only the input itself. Despite recent progress in pretrained image editing, such models often underperform on manga because they are trained predominantly on natural-image data, while re-training or fine-tuning them on manga is costly and raises copyright concerns. Many manga image editing tasks encountered in practice are structure-preserving, requiring local details to be modified while the input's global composition is retained. To support this common editing setting, our method corrects the early editing trajectory by anchoring it to an empty-prompt reconstruction trajectory. Experiments indicate improved performance in the main text-removal setting, while qualitative examples suggest better composition preservation in screentone synthesis. With FLUX.1 Kontext on an RTX A6000, the method incurs 11\% runtime overhead and 0.1\% peak-memory overhead; an additional runtime measurement with Qwen Image Edit 2509 on an NVIDIA H200 shows only a 0.1\% increase.
\end{abstract}


\section{Introduction}
Manga image editing is an important task in visual content creation.
Manga, a form of Japanese comics, has attracted a broad global audience, creating demand for a variety of image editing techniques.
For instance, when manga is localized into other languages, text embedded in the image must be removed seamlessly as a preprocessing step.
Moreover, a method that directly transforms line art into a completed manga image with screentones would provide practical assistance to creators.

Recently, the computer vision community has seen rapid progress in image editing methods built on large pretrained image generation models~\cite{huang2025diffusion}.
In particular, methods based on Rectified Flow~\cite{liu2023flow,lipman2023flow}, which enable high-quality image generation, have become the dominant paradigm~\cite{kulikov2025flowedit, deng2025fireflow, rout2025semantic,wang2025taming,patel2025flowchef,hu2024latent}.
These methods perform editing by re-generating an image that follows a target prompt from an input image associated with a source prompt using a pretrained image generation model.
More recently, models such as Qwen Image Edit and FLUX.1 Kontext, which are trained explicitly for image editing (hereafter, \emph{image editing models}), have emerged~\cite{wu2025qwen,labs2025flux,liu2025step1x,team2025longcat}.
These pretrained image editing models can be driven by direct editing instructions (e.g., ``Remove text.'' for removing text from an image), enabling more intuitive and flexible editing than methods based on generic image generation models.

\begin{figure}[t]
    \centering
    \begin{subfigure}[b]{0.32\linewidth}
        \centering
        \includegraphics[width=\linewidth]{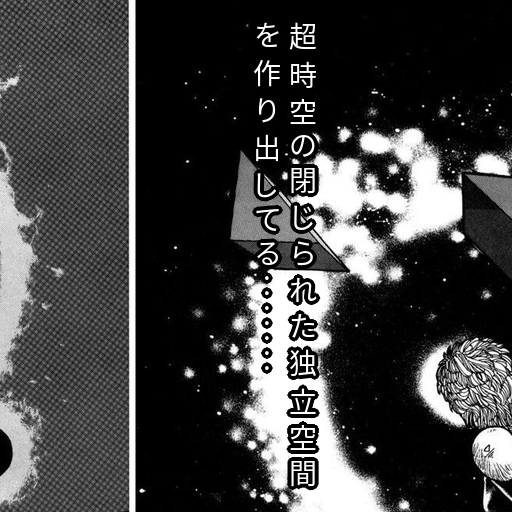}
        \caption{Input}
        \label{fig:noprompt_input}
    \end{subfigure}
    \hfill
    \begin{subfigure}[b]{0.32\linewidth}
        \centering
        \includegraphics[width=\linewidth]{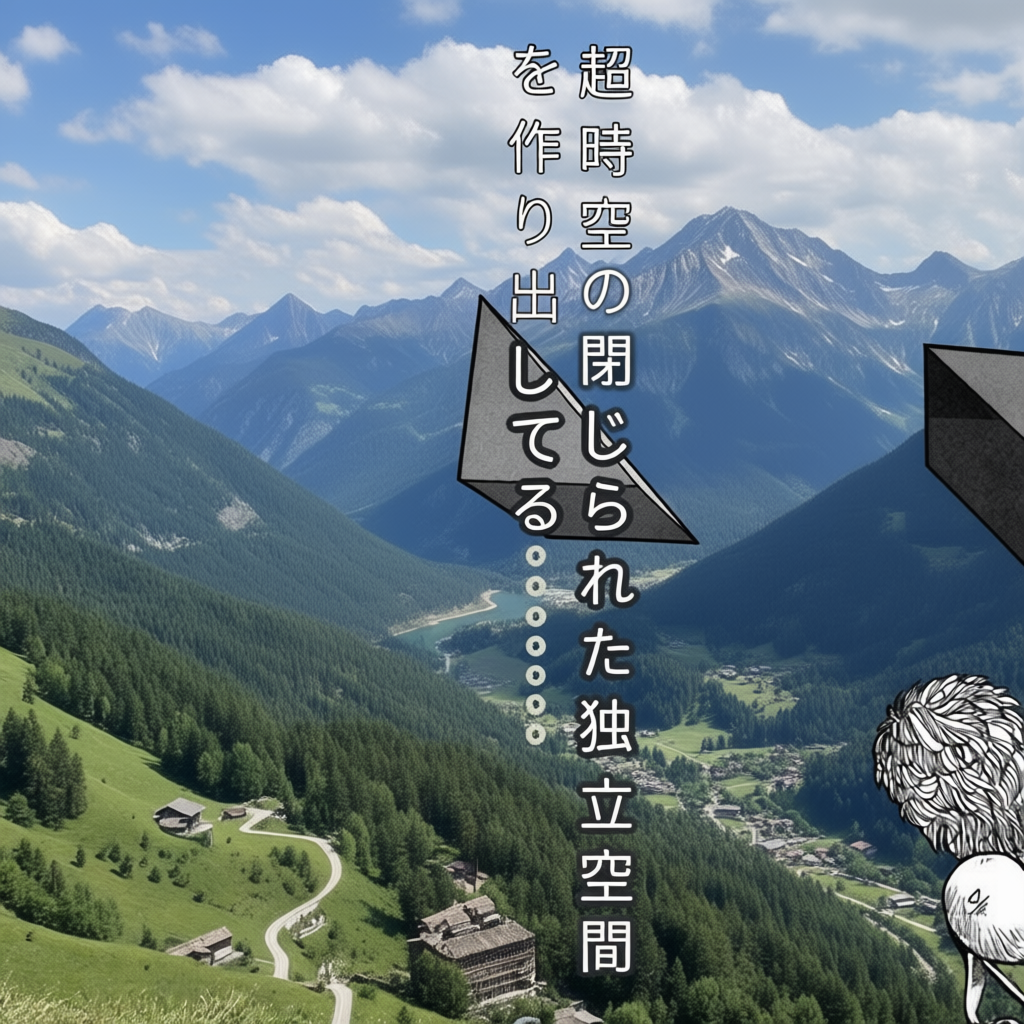}
        \caption{Baseline}
        \label{fig:noprompt_baseline}
    \end{subfigure}
    \hfill
    \begin{subfigure}[b]{0.32\linewidth}
        \centering
        \includegraphics[width=\linewidth]{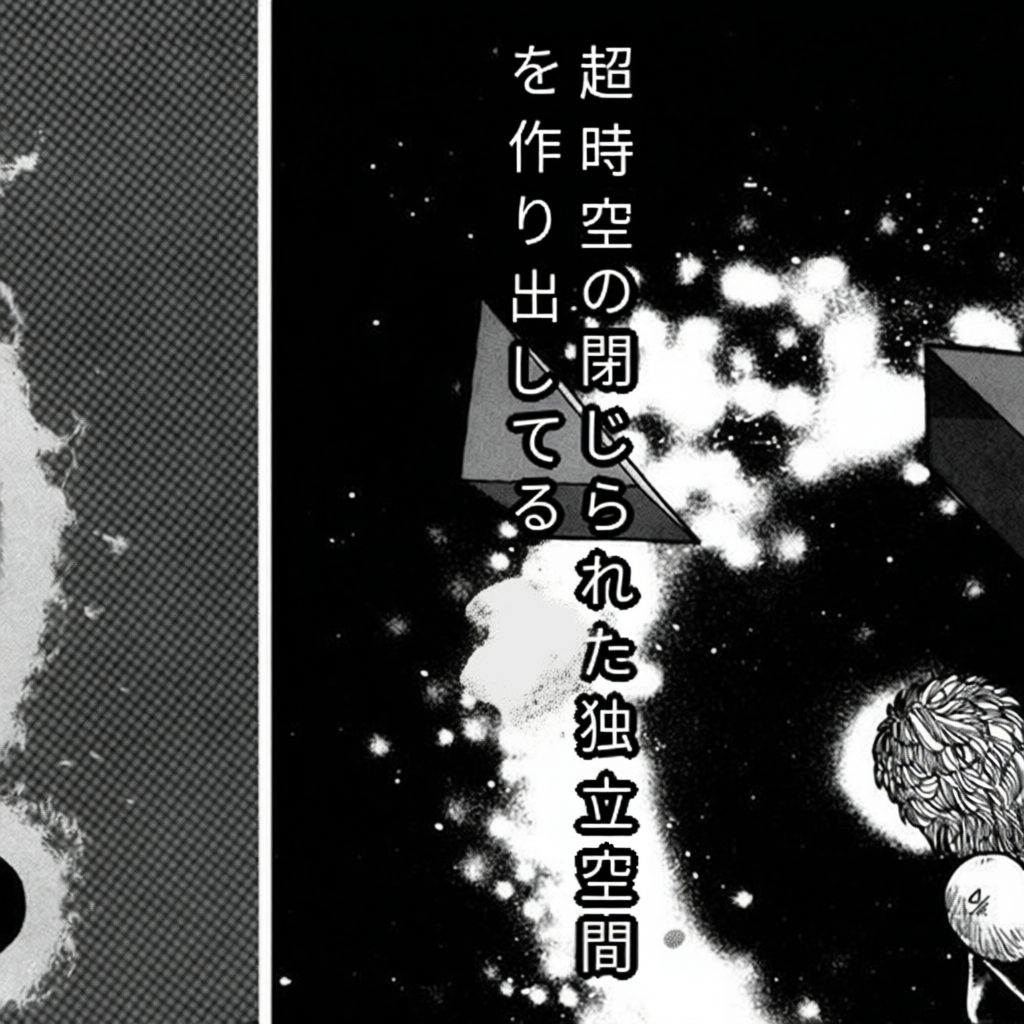}
        \caption{Ours}
        \label{fig:noprompt_proposed}
    \end{subfigure}
    \caption{Editing results with an empty prompt. \copyright Kato Masaki}
    \label{fig:noprompt}
\end{figure}

However, these pretrained image editing models often do not work well when directly applied to manga image editing.
For example, when we feed the manga image in Fig.~\ref{fig:noprompt_input} to the model and provide an ``empty prompt'' that specifies no edit, the model still transforms the image into a natural-image-like result as shown in Fig.~\ref{fig:noprompt_baseline}.
This failure to preserve the input even when no edit is requested suggests that the same domain bias can distort the global composition during local manga edits.
A plausible reason is that the data used to train these image editing models consist mostly of natural images, while manga images are relatively scarce.

A straightforward remedy would be to collect manga images and fine-tune an image editing model, but this is often impractical.
Image editing models such as Qwen Image Edit and FLUX.1 Kontext are large and require substantial computational resources for training.
Even if efficient adaptation methods such as LoRA~\cite{hu2022lora} are used, collecting manga images and using them for training still raises copyright concerns.

\begin{figure}[t]
    \centering
    \setlength{\fboxrule}{0.5pt}
    \setlength{\fboxsep}{0pt}

    \begin{subfigure}[t]{0.49\linewidth}
        \centering
        \makebox[0.48\linewidth][c]{\fbox{\includegraphics[height=0.46\linewidth]{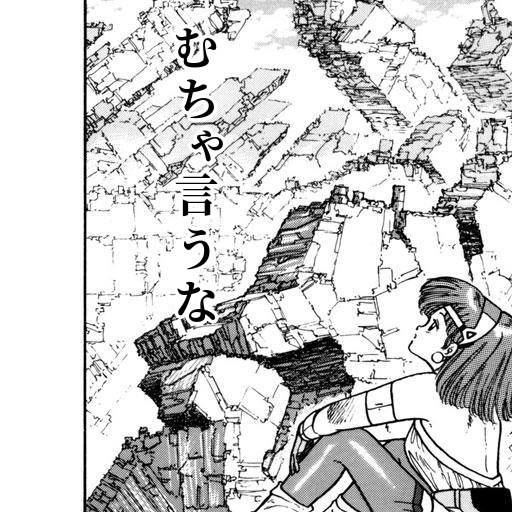}}}
        \hfill
        \makebox[0.48\linewidth][c]{\fbox{\includegraphics[height=0.46\linewidth]{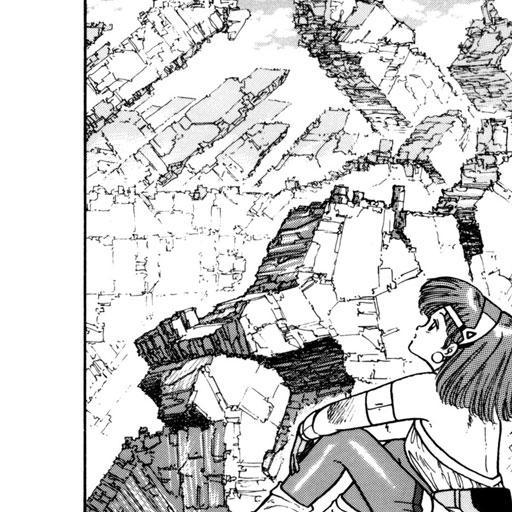}}}
        \caption{Text removal}
        \label{fig:data_text_removal}
    \end{subfigure}
    \hfill
    \begin{subfigure}[t]{0.49\linewidth}
        \centering
        \makebox[0.48\linewidth][c]{\fbox{\includegraphics[height=0.46\linewidth]{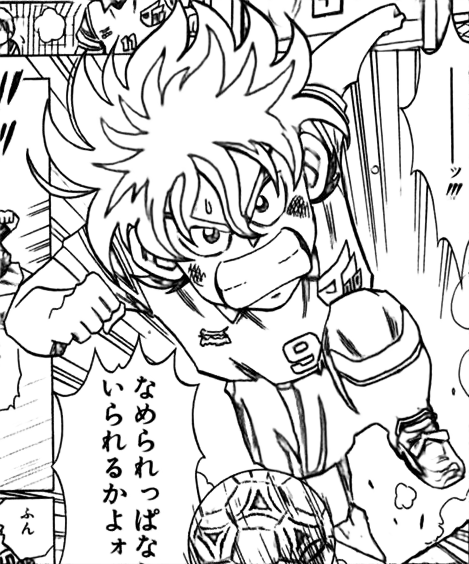}}}
        \hfill
        \makebox[0.48\linewidth][c]{\fbox{\includegraphics[height=0.46\linewidth]{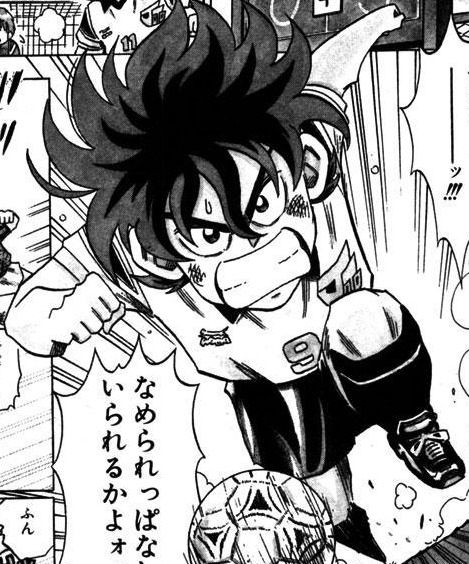}}}
        \caption{Screentone synthesis}
        \label{fig:data_screentone_synth}
    \end{subfigure}

    \caption{Examples of manga image editing. In each pair, the left image is the input and the right image is the ground truth. \copyright Kato Masaki \copyright Yabuno Tenya, Watanabe Tatsuya}
    \label{fig:data}
\end{figure}

To address this issue, we propose an input-specific correction of the inference trajectory that requires neither parameter updates nor external manga data.
Our approach builds on three observations.
First, Rectified Flow-based image editing models, including Qwen Image Edit and FLUX.1 Kontext, synthesize an edited image from noise while conditioning on the input image, rather than directly modifying its pixels; their editing behavior can therefore be adjusted by intervening in the generation trajectory at inference time.
Second, many practical manga editing tasks, including text removal and screentone synthesis, are structure-preserving: the output should retain the input's global composition while modifying only local details (Fig.~\ref{fig:data}).
Here, we use \emph{global composition} to refer collectively to the spatial layout and the structural line art that defines characters, objects, and panels; local details include elements such as text and screentone patterns.
Third, diffusion/flow models determine global structure mainly in the early timesteps and refine details in later timesteps~\cite{qian2024boosting,wang2023diffusion,ma2025nami}.
Together, these observations motivate correcting only the first few steps of the editing trajectory, where global composition is primarily determined.
Specifically, at each corrected step, we use the model's empty-prompt prediction to construct a correction target whose predicted one-step endpoint is the input image and anchor the editing trajectory toward this target.
The resulting correction is lightweight and practical: it introduces no trainable module, leaves the pretrained architecture unchanged, and can be incorporated into existing Rectified Flow-based image editing models through a sampler-level modification.

We evaluate the proposed method on two complementary structure-preserving manga editing tasks: text removal, which removes existing content, and screentone synthesis, which adds local visual patterns to line art.
Because the method assumes only a Rectified Flow-based editing process and does not depend on a particular network architecture, we test it across several Qwen and FLUX backbones.
For text removal, quantitative and qualitative improvements are observed in the main Qwen Image Edit 2509 setting.
Results on the other backbones further show that the same correction is applicable without model-specific training and can improve a range of text-removal metrics.
For screentone synthesis, qualitative comparisons in the supplementary material show that the proposed correction better preserves the input composition.

We also evaluate computational efficiency.
The edit-prompt and empty-prompt predictions are computed together in a single batched model call at each corrected step, so the proposed correction requires neither parameter updates nor an additional \emph{sequential} neural-network pass.
Nevertheless, the batched prediction increases the total computation and minibatch memory footprint; we therefore measure both runtime and peak memory for FLUX.1 Kontext on an A6000 GPU and additionally measure runtime for Qwen Image Edit 2509 on an H200 GPU.
For the A6000/FLUX.1 Kontext configuration, the runtime and peak-memory overhead are 11\% and 0.1\%, respectively, while the additional H200/Qwen Image Edit 2509 measurement shows a 0.1\% runtime increase (Sec.~\ref{sec:runtime}).

Finally, the input anchor makes the method particularly well suited to local edits that preserve global composition.
One current limitation is that the correction is not designed for edits requiring substantial changes in shape, pose, layout, or semantics.
Extending the method with an adaptive anchoring mechanism for such edits is an important direction for future work.

Our contributions are summarized as follows:
\begin{itemize}
    \item To the best of our knowledge, this is the first inference-time adaptation method for image editing models.
    \item We introduce a lightweight, training-free trajectory correction that uses empty-prompt reconstruction as an input-specific anchor for preserving global composition, without training or architectural changes.
    \item Experiments indicate improved performance in the main text-removal setting and better composition preservation in qualitative screentone-synthesis results, with small practical overhead.
\end{itemize}

\section{Related Work}
\subsection{Test-Time Adaptation}
Test-time adaptation, which aims to handle distribution shifts between training and inference data, has been studied extensively in image classification~\cite{liang2025comprehensive}.
A representative line of work updates a subset of model parameters, such as those in batch normalization layers, at inference time using the predicted class probabilities~\cite{wang2021tent,niu2022efficient}.
More recently, methods have also been proposed to update either the input image or model parameters at inference time by leveraging feedback or guidance from image generation models~\cite{prabhudesai2023diffusion,tsai2024gda,guo2025everything}.
However, these methods typically suffer from one or more of the following limitations: architectural constraints, task specificity to image classification, or substantial additional computational cost at inference time.

\citet{chen2025test} proposed an efficient method for large pretrained segmentation models that updates image embeddings (latent representations).
However, their method, as well as the classification-oriented approaches above, relies on classification or segmentation outputs and objectives that are unavailable in an image editing model; it therefore cannot be directly applied to our setting.

In contrast to these prior works, we present the first test-time adaptation method for recently emerged image editing models.
Instead of defining an adaptation objective from a classifier or segmenter output, our method derives a correction directly from the reconstruction behavior of the image editing model itself.

\subsection{Inversion and Trajectory-Based Editing}
Several recent methods perform image inversion or editing by modifying the inference dynamics or intermediate representations of flow-based generative models.
RF-Inversion~\cite{rout2025semantic} derives an inversion vector field through dynamic optimal control and establishes its equivalence to a rectified stochastic differential equation, while FireFlow~\cite{deng2025fireflow} accelerates inversion by reusing the preceding step's midpoint velocity.
RF-Solver~\cite{wang2025taming} reduces numerical errors in sampling and inversion, and RF-Edit uses inversion features to preserve source structure during image and video editing.
\citet{hu2024latent} edit transformer-based flow-matching models through a controllable $u$-space and localized text prompts.
FlowEdit~\cite{kulikov2025flowedit} constructs an inversion-free ODE that directly maps between source and target distributions, whereas FlowChef~\cite{patel2025flowchef} steers the denoising vector field during inference for controlled generation.
Whereas these methods adapt flow-based generation models for editing by modifying inversion, inference dynamics, or intermediate features, we correct the trajectory of an image editing model that already accepts an input image and an edit instruction.
Specifically, to preserve global composition during local editing, we derive an input-specific anchor from the same model's empty-prompt reconstruction trajectory and use it to correct the early edit trajectory.
We empirically compare with representative inversion and trajectory-based methods in Sec.~\ref{sec:text_comparison}.

\subsection{Manga Image Editing}
A wide range of tasks have been studied in manga image editing.
Representative examples include automatic colorization~\cite{golyadkin2025closing,hensman2017cgan,furusawa2017comicolorization,shimizu2021painting,qu2006manga,kataoka2017automatic,sato2014reference}, text removal~\cite{ko2020sickzil,xie2021seamless}, screentone synthesis~\cite{tsubota2019synthesis,lin2024sketch2manga}, and retargeting~\cite{matsui2011interactive,xie2025screentone}.
Unlike these task-specific approaches, our method adapts a pretrained image editing model to each input at inference time without training a dedicated model for the target task.
It therefore avoids collecting a task-specific manga training set and can change the requested local edit through the instruction prompt.
Our experiments evaluate this task-flexible adaptation on text removal and screentone synthesis, two complementary tasks for which quantitative evaluation can be constructed using manga with documented research-use permission.

\section{Method}

\begin{algorithm}[t]
    \caption{Inference-time trajectory optimization}
    \label{alg:inference}
    \begin{algorithmic}[1]
    \Require Pretrained Rectified-flow model $\bm{v}_{\theta}$, input image $\bm{X}_{\mathrm{in}}$, noise $\bm{Z}_0$, and edit prompt $c$
    \For{$i \in \{0,\cdots,N-1\}$}
        \State $\bm{v} \leftarrow \bm{v}_{\theta}(\bm{Z}_{t_i}, t_i, c, \bm{X}_{\mathrm{in}})$
        \If{\textcolor{red}{$i<M$}}
            \State {\color{red}$\bm{u} \leftarrow \bm{v}_{\theta}(\bm{Z}_{t_i}, t_i, \emptyset, \bm{X}_{\mathrm{in}})$}
            \State {\color{red}$\bm{Z}^*_{t_i} \leftarrow \bm{X}_{\mathrm{in}} - (t_N - t_i)\bm{u}$}
            \State {\color{red}$\bm{Z}_{t_i} \leftarrow (1-\alpha) \bm{Z}_{t_i} + \alpha\bm{Z}^*_{t_i}$}
        \EndIf
        \State $\bm{Z}_{t_{i+1}} \leftarrow \bm{Z}_{t_i} + (t_{i+1}-t_i)\bm{v}$
    \EndFor
    \State \Return $\bm{Z}_{t_N}$
    \end{algorithmic}
\end{algorithm}

\subsection{Preliminaries: Generation Process of Image Editing Models}
Rectified Flow-based image editing models such as Qwen Image Edit and FLUX.1 Kontext do not edit the input image directly.
Instead, conditioned on the input image, they generate an edited image from noise.
This generation process can be written as the following ordinary differential equation:
\begin{equation}
    d\bm{Z}_t=\bm{v}_{\theta}(\bm{Z}_t, t, c, \bm{X}_{\mathrm{in}})dt,
\end{equation}
where $t\in[0,1]$ denotes time, $c$ is the editing prompt, $\bm{Z}_0 \sim \mathcal{N}(\bm{0}, \bm{I})$ is Gaussian noise, and $\bm{Z}_1 = \bm{X}_{\mathrm{edit}}$ is the edited image.
Here, $\bm{v}_\theta$ denotes the velocity predicted by the pretrained image editing model.
In practice, time $t$ is discretized into $N$ timesteps $t=(t_0,\cdots,t_{N})$, and generation is performed with an Euler solver:
\begin{equation}
    \bm{Z}_{t_{i+1}}=\bm{Z}_{t_{i}}+(t_{i+1}-t_{i})\bm{v}_{\theta}(\bm{Z}_{t_i}, t_i, c, \bm{X}_{\mathrm{in}}).
\end{equation}
Starting from $t_0=0$ and ending at $t_N=1$, we iterate the above update for $i\in\{0,\cdots,N-1\}$ to obtain the edited image $\bm{Z}_1 (= \bm{X}_{\mathrm{edit}})$.

\subsection{Correcting the Generation Trajectory}
The key difficulty is that when $\bm{X}_{\mathrm{in}}$ lies outside the training distribution, as is often the case for manga images, the predicted velocity $\bm{v}_{\theta}(\bm{Z}_{t_i}, t_i, c, \bm{X}_{\mathrm{in}})$ becomes inaccurate, leading to trajectories and outputs that are unsuitable for manga images.
Our method addresses this issue by optimizing the generation trajectory.

A key property of manga editing tasks such as text removal and screentone synthesis is that the global composition of the input image $\bm{X}_{\mathrm{in}}$ and the edited image $\bm{X}_{\mathrm{edit}}$ is expected to remain largely unchanged.
In other words, the edited output should preserve the input's global composition while changing only local details.
Given the observation that diffusion/flow models generate global structure in early timesteps and fine details in later timesteps~\cite{qian2024boosting,wang2023diffusion,ma2025nami}, it is desirable for the trajectory of the edited image to stay close to that of the input image during the first $M (<N)$ steps, $t=(t_0,\cdots,t_M)$.
We therefore compute a trajectory that reconstructs the input image under an empty prompt and use it to correct the trajectory of the edited image.

More specifically, at timestep $t_i$, let
$\bm{u}_i:=\bm{v}_{\theta}(\bm{Z}_{t_i},t_i,\emptyset,\bm{X}_{\mathrm{in}})$
be the empty-prompt velocity evaluated at the current state.
Under the one-step approximation, we freeze this velocity locally and define the endpoint prediction for a candidate state $\bm{Z}$ as
\begin{equation}
    \hat{\bm{Z}}_1(\bm{Z})=\bm{Z}+(t_N-t_i)\bm{u}_i.
\end{equation}
We then define the correction target by minimizing the discrepancy between this predicted endpoint and $\bm{X}_{\mathrm{in}}$:
\begin{align}
    \bm{Z}^*_{t_i}
    &=\arg\min_{\bm{Z}}
    \left\|\bm{X}_{\mathrm{in}}-\hat{\bm{Z}}_1(\bm{Z})\right\|_2^2
    \label{eq:optimization} \\
    &=\bm{X}_{\mathrm{in}}-(t_N-t_i)\bm{u}_i .
    \label{eq:opt_latent}
\end{align}
Using this result, we correct the original trajectory $\bm{Z}_{t_i}$ as follows:
\begin{equation}
    \bm{Z}_{t_i} \leftarrow (1-\alpha) \bm{Z}_{t_i} + \alpha \bm{Z}^*_{t_i}, \label{eq:fixation}
\end{equation}
where $\alpha$ is a hyperparameter that controls the correction strength.

\subsection{Algorithm and Computational Cost}\label{sec:alg}
The full procedure is summarized in Algorithm~\ref{alg:inference}.
The red lines indicate the additional computation introduced by our method.
For the first $M$ steps, the prediction with the edit prompt $c$ (line 2) and that with the empty prompt $\emptyset$ (line 4) can be executed in one batched model call by concatenating the two conditions.
This avoids an additional \emph{sequential} model pass, but still increases the amount of computation and the minibatch memory footprint.
The interpolation in lines 5 and 6 adds only simple matrix operations.
This lightweight correction is easy to add to an existing sampler and requires neither modification nor retraining of the underlying model.
We report runtime and peak memory for FLUX.1 Kontext on an A6000 GPU together with an additional runtime measurement for Qwen Image Edit 2509 on an H200 GPU in Sec.~\ref{sec:runtime}.

\section{Mathematical Interpretation}
\label{sec:math_interpretation}
This section analyzes the proposed correction while holding the velocity predictions computed at timestep $t_i$ fixed, matching how Algorithm~\ref{alg:inference} constructs and applies the correction target.
Sec.~\ref{sec:local_reconstruction_effect} establishes the reconstruction guarantee of the interpolation, Sec.~\ref{sec:why_empty_prompt} explains the choice of the empty prompt, and Sec.~\ref{sec:flowchef_connection} clarifies the mathematical relationship to FlowChef.

\subsection{Reconstruction Guarantee of the Interpolation}
\label{sec:local_reconstruction_effect}
This subsection shows that the interpolation correction reduces the discrepancy between the input image and the endpoint predicted from the empty-prompt velocity.
At timestep $t_i$, define
\begin{equation}
\mathbf{u}_i := \mathbf{v}_\theta(\mathbf{Z}_{t_i},t_i,\emptyset,\mathbf{X}_{\rm in}),
\;
\Delta_i := t_N-t_i .
\end{equation}
Here, $\mathbf{u}_i$ is the empty-prompt velocity and $\Delta_i$ is the remaining time to the endpoint.
Holding $\mathbf{u}_i$ fixed, a candidate current state $\mathbf{Z}$ predicts the endpoint $\mathbf{Z}+\Delta_i\mathbf{u}_i$.
The correction target
\begin{equation}
\mathbf{Z}_{t_i}^{*}
=
\mathbf{X}_{\rm in}-\Delta_i\mathbf{u}_i
\end{equation}
is therefore the unique state whose predicted empty-prompt endpoint equals the input image.
For the current state, define the endpoint reconstruction residual as
\begin{equation}
\mathbf{r}_i
:=
\mathbf{Z}_{t_i}+\Delta_i\mathbf{u}_i-\mathbf{X}_{\rm in}.
\end{equation}
The target can equivalently be written as $\mathbf{Z}_{t_i}^{*}=\mathbf{Z}_{t_i}-\mathbf{r}_i$: it removes the entire frozen-velocity residual.
Algorithm~\ref{alg:inference} moves only a fraction $\alpha$ toward this target,
\begin{equation}
\mathbf{Z}_{t_i}^{+}
:=
(1-\alpha)\mathbf{Z}_{t_i}
+
\alpha\mathbf{Z}_{t_i}^{*}.
\end{equation}

\noindent\textbf{Proposition 1 (Exact local reconstruction contraction).}
Under the frozen velocity $\mathbf{u}_i$, the corrected state is
\begin{equation}
\mathbf{Z}_{t_i}^{+}=\mathbf{Z}_{t_i}-\alpha\mathbf{r}_i,
\end{equation}
and the residual after correction satisfies
\begin{equation}
\mathbf{r}_i^{+}
:=
\mathbf{Z}_{t_i}^{+}+\Delta_i\mathbf{u}_i-\mathbf{X}_{\rm in}
=
(1-\alpha)\mathbf{r}_i .
\end{equation}
Consequently,
\begin{equation}
\|\mathbf{r}_i^{+}\|_2
=
|1-\alpha|\,\|\mathbf{r}_i\|_2,
\qquad
\|\mathbf{r}_i^{+}\|_2^2
=
(1-\alpha)^2\|\mathbf{r}_i\|_2^2 .
\end{equation}

\noindent\textbf{Proof.}
Substituting $\mathbf{Z}_{t_i}^{*}=\mathbf{Z}_{t_i}-\mathbf{r}_i$ into the interpolation update gives
$\mathbf{Z}_{t_i}^{+}=\mathbf{Z}_{t_i}-\alpha\mathbf{r}_i$.
Adding $\Delta_i\mathbf{u}_i-\mathbf{X}_{\rm in}$ to both sides yields
$\mathbf{r}_i^{+}=(1-\alpha)\mathbf{r}_i$, from which the norm identities follow.
\hfill$\square$

Although later timesteps use newly predicted velocities and final-task performance lies outside this analysis, Proposition 1 guarantees the intended immediate effect: for $0<\alpha<1$, the current empty-prompt endpoint estimate moves strictly toward the input, and $\alpha=1$ eliminates the frozen-velocity residual.
This result requires neither a network Jacobian nor a Lipschitz assumption.

\subsection{Why the Empty Prompt Helps}
\label{sec:why_empty_prompt}
To clarify why the correction target is constructed with the empty prompt, we compare the proposed correction with a hypothetical variant that uses the edit prompt $c$ instead.
Both variants use the edit-prompt velocity for the actual Euler update; they differ only in the prompt used to construct the correction target.
Let $h_i:=t_{i+1}-t_i$ denote the Euler step size.
Let the edit-prompt velocity and its difference from the empty-prompt velocity be
\begin{equation}
\mathbf{v}_i^{(c)}
:=
\mathbf{v}_\theta(\mathbf{Z}_{t_i},t_i,c,\mathbf{X}_{\rm in}),
\qquad
\boldsymbol{\delta}_i
:=
\mathbf{v}_i^{(c)}-\mathbf{u}_i .
\end{equation}
Thus, $\mathbf{v}_i^{(c)}=\mathbf{u}_i+\boldsymbol{\delta}_i$.

\noindent\textbf{Proposition 2 (Effect of the prompt used for target construction).}
The proposed empty-prompt target and the hypothetical edit-prompt target are, respectively,
\begin{equation}
\mathbf{Z}_{t_i}^{*,(\emptyset)}
=
\mathbf{X}_{\rm in}-\Delta_i\mathbf{u}_i,
\qquad
\mathbf{Z}_{t_i}^{*,(c)}
=
\mathbf{X}_{\rm in}-\Delta_i(\mathbf{u}_i+\boldsymbol{\delta}_i).
\end{equation}
For either choice $p\in\{\emptyset,c\}$, one correction step followed by the Euler update is
\begin{equation}
\mathbf{Z}_{t_{i+1}}^{(p)}
=
(1-\alpha)\mathbf{Z}_{t_i}
+
\alpha\mathbf{Z}_{t_i}^{*,(p)}
+
h_i(\mathbf{u}_i+\boldsymbol{\delta}_i).
\end{equation}
The target $\mathbf{Z}_{t_i}^{*,(p)}$ is the only term in this update that differs between the two variants.
Substituting the two targets yields
\begin{equation}
\begin{aligned}
\mathbf{Z}_{t_{i+1}}^{(\emptyset)}
&=
(1-\alpha)\mathbf{Z}_{t_i}
+\alpha\mathbf{X}_{\rm in}
+(h_i-\alpha\Delta_i)\mathbf{u}_i
+h_i\boldsymbol{\delta}_i, \\
\mathbf{Z}_{t_{i+1}}^{(c)}
&=
(1-\alpha)\mathbf{Z}_{t_i}
+\alpha\mathbf{X}_{\rm in}
+(h_i-\alpha\Delta_i)\mathbf{u}_i
+(h_i-\alpha\Delta_i)\boldsymbol{\delta}_i .
\end{aligned}
\end{equation}
Their difference is therefore
\begin{equation}
\mathbf{Z}_{t_{i+1}}^{(c)}
-
\mathbf{Z}_{t_{i+1}}^{(\emptyset)}
=
-\alpha\Delta_i\boldsymbol{\delta}_i .
\end{equation}

\noindent\textbf{Proof.}
Substituting each target into the correction-plus-Euler update above gives the two stated expressions.
Subtracting them gives the final identity.
\hfill$\square$

The first three terms in the two next-state expressions are identical; only the coefficient of $\boldsymbol{\delta}_i$ differs.
Because $\boldsymbol{\delta}_i$ is the change in velocity caused by replacing the empty prompt with the edit prompt, $h_i\boldsymbol{\delta}_i$ is the part added by the original Euler step in response to the edit instruction.
The proposed empty-prompt target keeps its coefficient at $h_i$, so the proposed correction does not implicitly rescale this response.
In contrast, the edit-prompt target introduces $-\alpha\Delta_i\boldsymbol{\delta}_i$ through the correction, changing the coefficient from $h_i$ to $h_i-\alpha\Delta_i$.
The coefficient is reduced, becomes zero when $\alpha\Delta_i=h_i$, and becomes negative when $\alpha\Delta_i>h_i$.
This interference is most relevant in the early steps, where $\Delta_i$ is largest and the proposed correction is applied.

\subsection{Connection to FlowChef}
\label{sec:flowchef_connection}
To clarify the mathematical relationship to FlowChef, we first compare the two update equations and then explain how empty-prompt reconstruction determines the proposed correction.

After gradient skipping, the FlowChef update~\cite{patel2025flowchef} takes the form
\begin{equation}
\mathbf{x}_{t-\Delta t}
=
\mathbf{x}_t
+
\underbrace{\Delta t\,\mathbf{u}_\theta(\mathbf{x}_t,t)}_{\text{base generation step}}
-
\underbrace{s\nabla_{\hat{\mathbf{x}}_0}L}_{\text{endpoint-objective steering}}.
\end{equation}
Here, $L$ is the endpoint objective and $s$ is its steering scale.
To express the proposed correction in the same form, define the squared endpoint reconstruction loss
\begin{equation}
\tilde L_i(\mathbf{Z})
:=
\|\mathbf{Z}+\Delta_i\mathbf{u}_i-\mathbf{X}_{\rm in}\|_2^2 .
\end{equation}
Since $\nabla_{\mathbf{Z}}\tilde L_i(\mathbf{Z}_{t_i})=2\mathbf{r}_i$, the interpolation in Proposition 1 is the exact gradient step
\begin{equation}
\mathbf{Z}_{t_i}^{+}
=
\mathbf{Z}_{t_i}
-\frac{\alpha}{2}\nabla_{\mathbf{Z}}\tilde L_i(\mathbf{Z}_{t_i}).
\end{equation}
Combining this correction with the Euler step, the first $M$ steps of the proposed sampler can be written as
\begin{equation}
\begin{aligned}
\mathbf{Z}_{t_{i+1}}
&=
\mathbf{Z}_{t_i}
+
\underbrace{h_i\mathbf{v}_\theta(\mathbf{Z}_{t_i},t_i,c,\mathbf{X}_{\rm in})}_{\text{base editing step}} \\
&\quad -
\underbrace{\frac{\alpha}{2}\nabla_{\mathbf{Z}}\tilde L_i(\mathbf{Z}_{t_i})}_{\text{reconstruction-based correction}},
\qquad i<M .
\end{aligned}
\end{equation}
Thus, both updates consist of a base vector-field step followed by a term that steers the predicted endpoint.
This shared algebraic form is the precise connection between the two methods.

FlowChef's endpoint-objective steering term and the proposed reconstruction-based correction have different origins.
FlowChef applies a general endpoint objective $L$ to an endpoint predicted from the vector field of an image generation model.
The proposed sampler instead uses an image editing model whose velocity prediction is explicitly conditioned on the input image $\mathbf{X}_{\rm in}$.
Within this model, the base step uses the edit-prompt velocity $\mathbf{v}_i^{(c)}$, whereas the loss defining the proposed correction uses the empty-prompt velocity $\mathbf{u}_i$.
This separation is what preserves the coefficient $h_i$ on $\boldsymbol{\delta}_i=\mathbf{v}_i^{(c)}-\mathbf{u}_i$ in Proposition 2.
Specifically, $\tilde L_i$ is the frozen empty-prompt reconstruction loss defined above.
Its unique minimizer is the input-specific target $\mathbf{Z}_{t_i}^{*}=\mathbf{X}_{\rm in}-\Delta_i\mathbf{u}_i$, which follows directly from the requirement to preserve the input structure.
Algorithm~\ref{alg:inference} implements the resulting update directly by interpolating toward this closed-form target during the first $M$ steps.
The gradient expression above is an exact algebraic interpretation of this interpolation.

Consequently, the shared update form provides a useful mathematical interpretation of the proposed correction.
Within this form, the paired edit- and empty-prompt predictions, the closed-form input-specific target, and the early-step schedule define a distinct construction tailored to structure-preserving editing.

\section{Experiments: Mask-Free Text Removal}

\begin{table}[t]
  \centering
  \caption{Evaluation of generation results under an empty prompt}
  \tablebodyfont
  \setlength{\tabcolsep}{0.5mm}
  \begin{tabular}{@{}lrrrrr@{}}
    \toprule
    \multirow{2}{*}{Method} & \multicolumn{3}{c}{Full image} & \multicolumn{2}{c}{Text region} \\
    \cmidrule(lr){2-4} \cmidrule(lr){5-6}
     & PSNR $\uparrow$ & SSIM $\uparrow$ & LPIPS $\downarrow$ & PSNR $\uparrow$ & SSIM $\uparrow$ \\
    \midrule
    Baseline & 11.84 & 48.02 & 41.13 & 8.71 & 28.90 \\
    Ours & \textbf{22.27} & \textbf{86.25} & \textbf{7.32} & \textbf{18.50} & \textbf{78.13} \\
    \bottomrule
  \end{tabular}
  \label{tab:noprompt}
\end{table}

\begin{table*}[t]
  \centering
  \caption{Quantitative comparison for text removal}
  \tablebodyfont
  \setlength{\tabcolsep}{0.5mm}
  \begin{tabular}{@{}llrrrrr@{}}
    \toprule
    \multirow{2}{*}{Method} & \multirow{2}{*}{Model} & \multicolumn{3}{c}{Full image} & \multicolumn{2}{c}{Text region} \\
    \cmidrule(lr){3-5} \cmidrule(lr){6-7}
     & & PSNR $\uparrow$ & SSIM $\uparrow$ & LPIPS $\downarrow$ & PSNR $\uparrow$ & SSIM $\uparrow$ \\
    \midrule
    Baseline & Qwen Image Edit 2509 & 19.62 & 72.76 & 7.16 & 19.07 & 60.77 \\
    Noise Inversion & Qwen Image Edit 2509 & 20.29 & 75.66 & 7.23 & 18.94 & 61.80 \\
    Ours & Qwen Image Edit 2509 & 22.07 & 80.61 & 6.00 & 20.04 & 64.15 \\
    FlowChef~\cite{patel2025flowchef} & Qwen Image 2512 & 13.97 & 52.11 & 39.68 & 5.68 & 8.42 \\
    FlowEdit~\cite{kulikov2025flowedit} & Qwen Image 2512 & 10.66 & 35.24 & 10.69 & 8.96 & 16.57 \\ \midrule
    Baseline & FLUX.1 Kontext [dev] & 26.42 & 90.47 & 6.60 & 21.00 & 66.97 \\
    Ours & FLUX.1 Kontext [dev] & \textbf{27.36} & 93.81 & \textbf{4.21} & \textbf{21.48} & \textbf{68.97} \\
    FlowChef~\cite{patel2025flowchef} & FLUX.1 [dev] & 16.91 & 75.44 & 19.43 & 4.83 & 5.54 \\
    FlowEdit~\cite{kulikov2025flowedit} & FLUX.1 [dev] & 14.48 & 68.98 & 28.18 & 5.59 & 7.59 \\
    FireFlow~\cite{deng2025fireflow} & FLUX.1 [dev] & 10.62 & 40.48 & 57.70 & 7.35 & 12.99 \\
    RF-Edit~\cite{wang2025taming} & FLUX.1 [dev] & 12.05 & 48.71 & 50.54 & 7.79 & 14.84 \\
    RF-Inversion~\cite{rout2025semantic} & FLUX.1 [dev] & 11.43 & 41.37 & 49.26 & 7.96 & 15.25 \\ \midrule
    SickZil-Machine~\cite{ko2020sickzil} & Finetuned CNNs & 24.94 & \textbf{95.30} & 5.05 & 16.16 & 48.8 \\
    \bottomrule
  \end{tabular}
  \label{tab:comparison_text_removal}
\end{table*}

\begin{figure*}[t]
    \centering
    \setlength{\fboxrule}{0.5pt}
    \setlength{\fboxsep}{0pt}

    \begin{minipage}[t]{0.13\textwidth}
        \centering
        \fbox{\includegraphics[width=\linewidth]{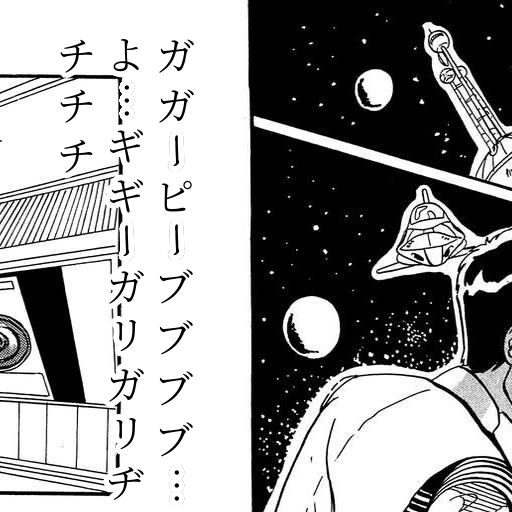}}\\[2pt]
        \fbox{\includegraphics[width=\linewidth]{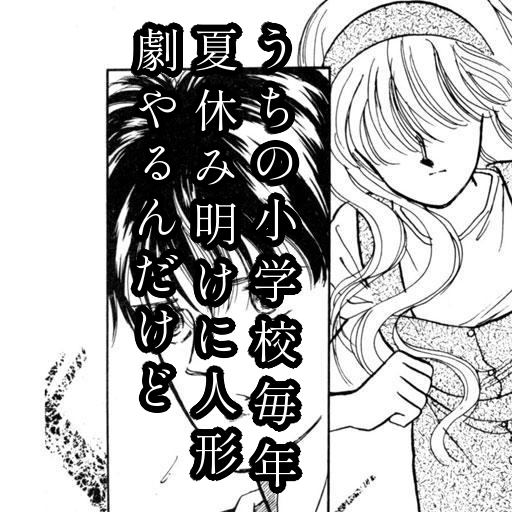}}\\[3pt]
        \small Input
    \end{minipage}
    \hfill
    \begin{minipage}[t]{0.13\textwidth}
        \centering
        \fbox{\includegraphics[width=\linewidth]{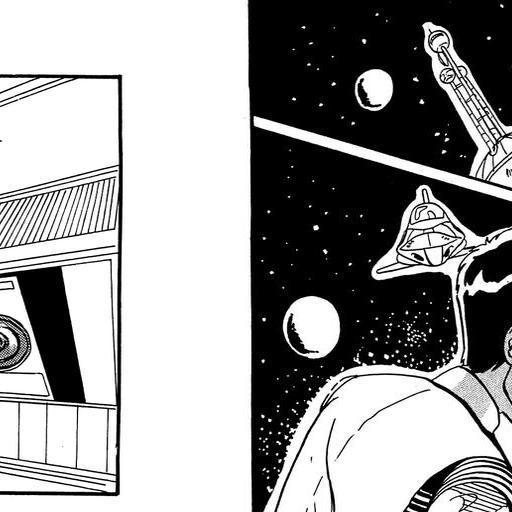}}\\[2pt]
        \fbox{\includegraphics[width=\linewidth]{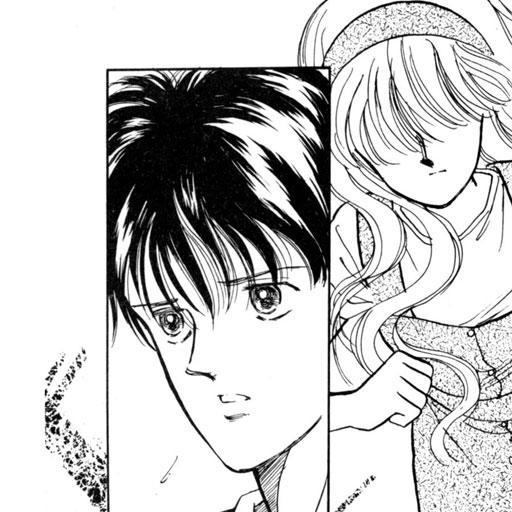}}\\[3pt]
        \small GT
    \end{minipage}
    \hfill
    \begin{minipage}[t]{0.13\textwidth}
        \centering
        \fbox{\includegraphics[width=\linewidth]{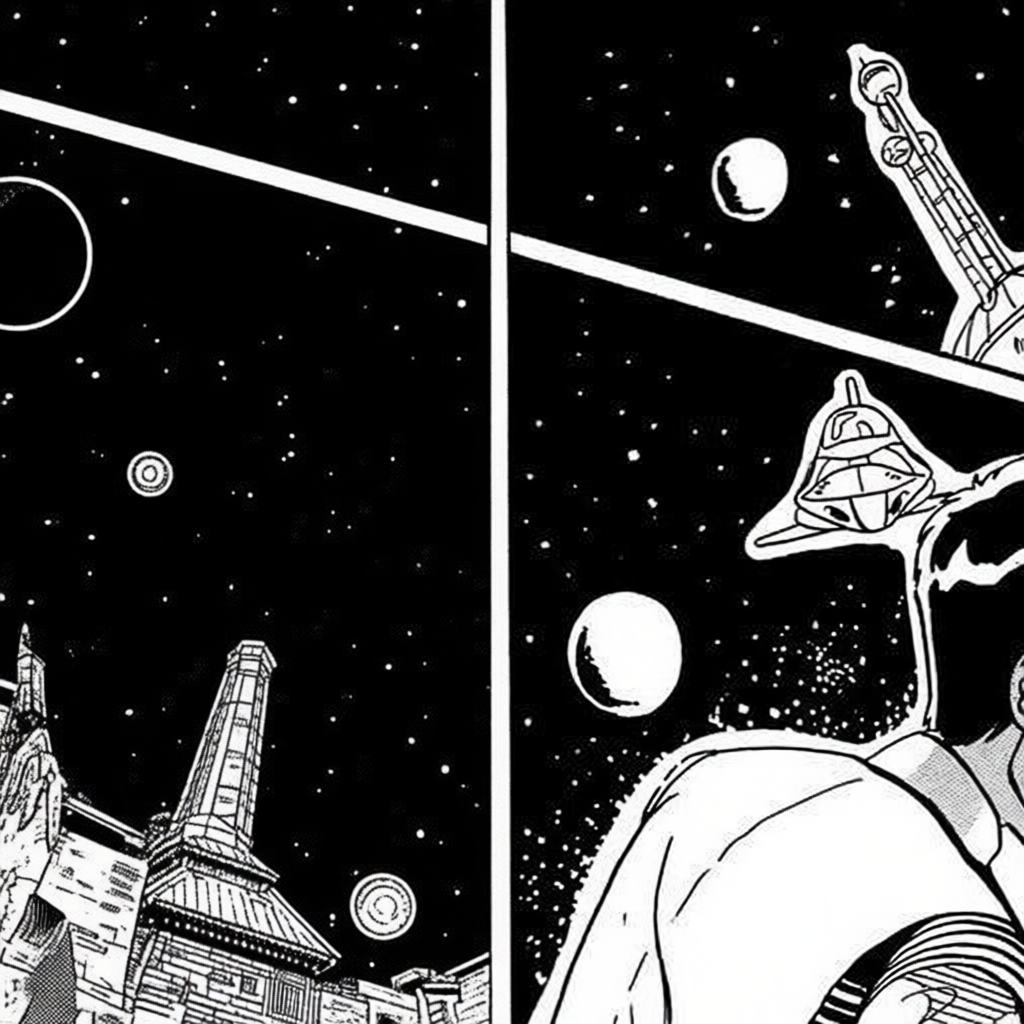}}\\[2pt]
        \fbox{\includegraphics[width=\linewidth]{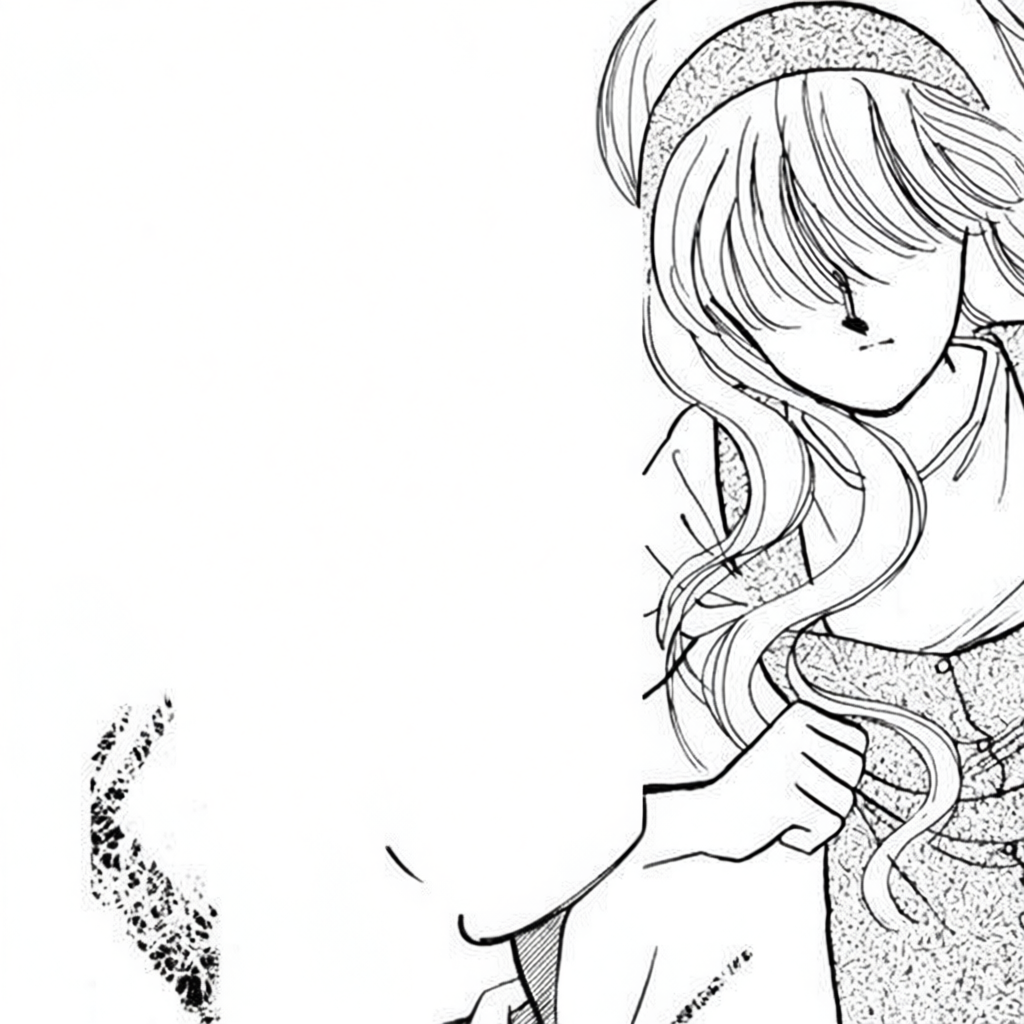}}\\[3pt]
        \small Baseline
    \end{minipage}
    \hfill
    \begin{minipage}[t]{0.13\textwidth}
        \centering
        \fbox{\includegraphics[width=\linewidth]{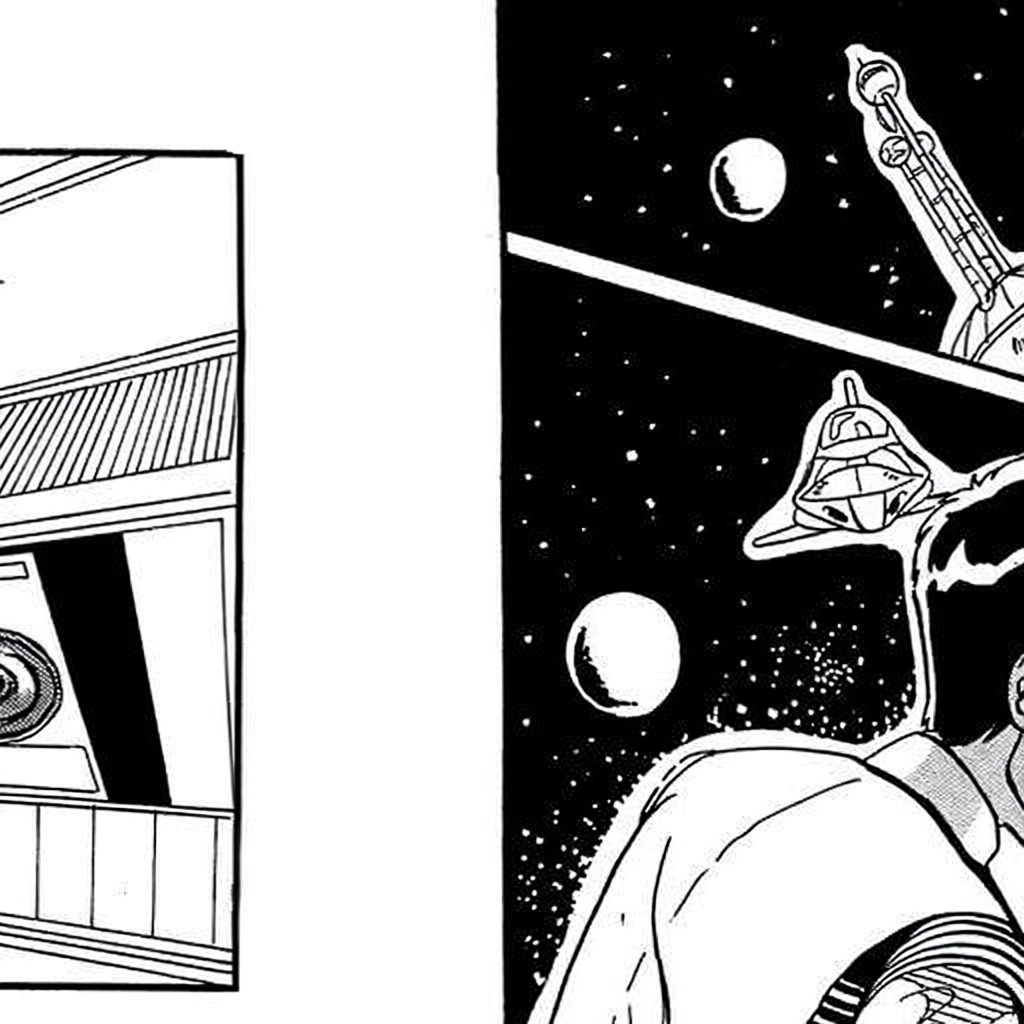}}\\[2pt]
        \fbox{\includegraphics[width=\linewidth]{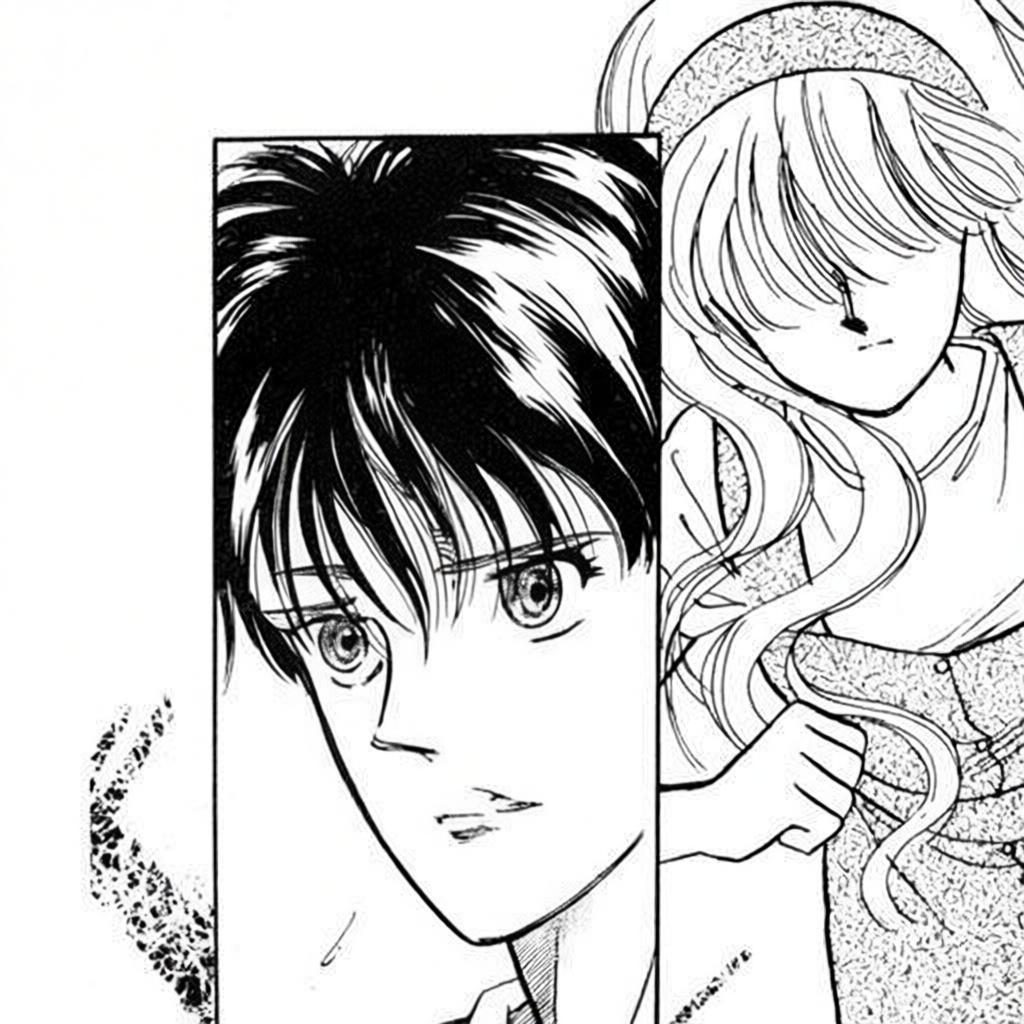}}\\[3pt]
        \small Ours
    \end{minipage}
    \hfill
    \begin{minipage}[t]{0.13\textwidth}
        \centering
        \fbox{\includegraphics[width=\linewidth]{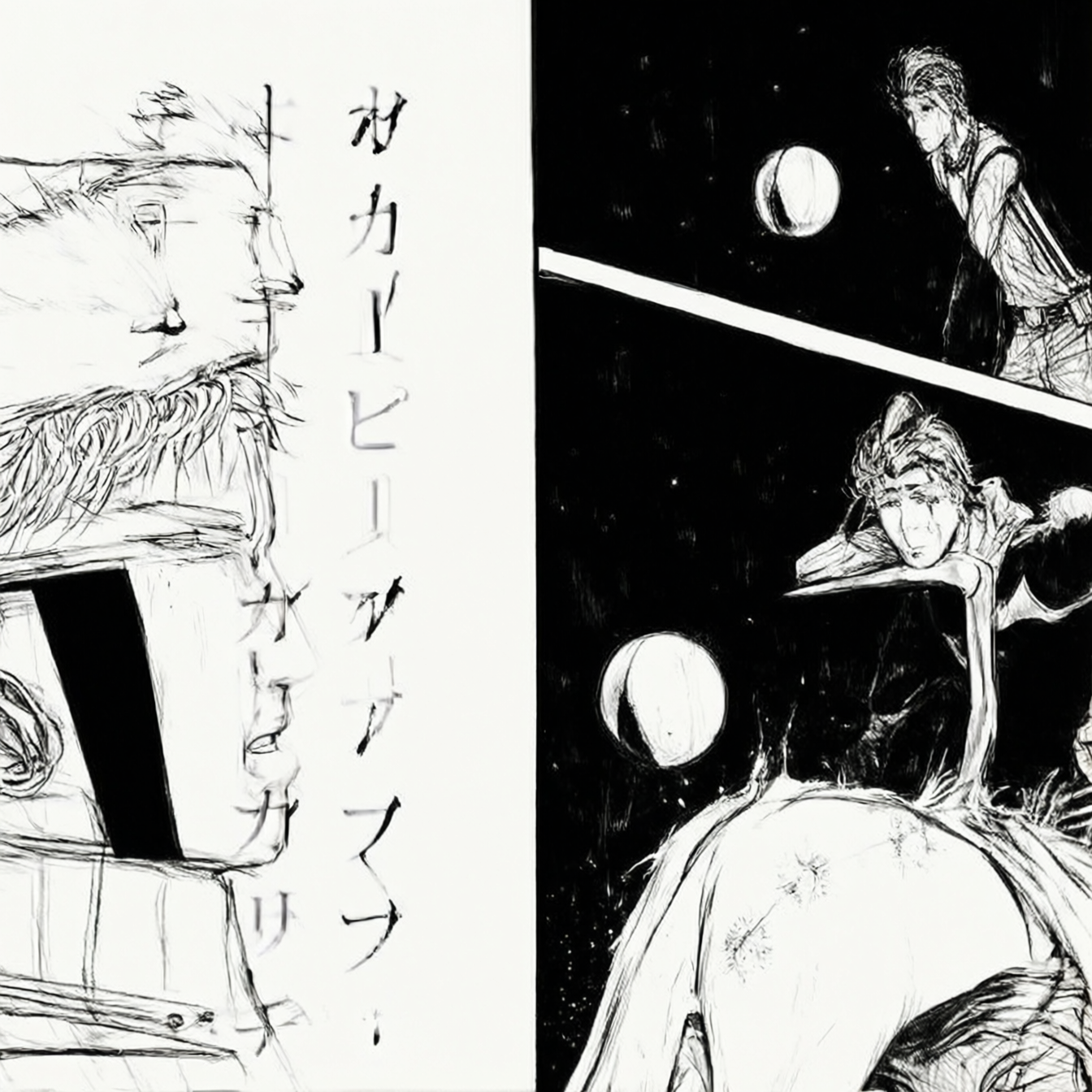}}\\[2pt]
        \fbox{\includegraphics[width=\linewidth]{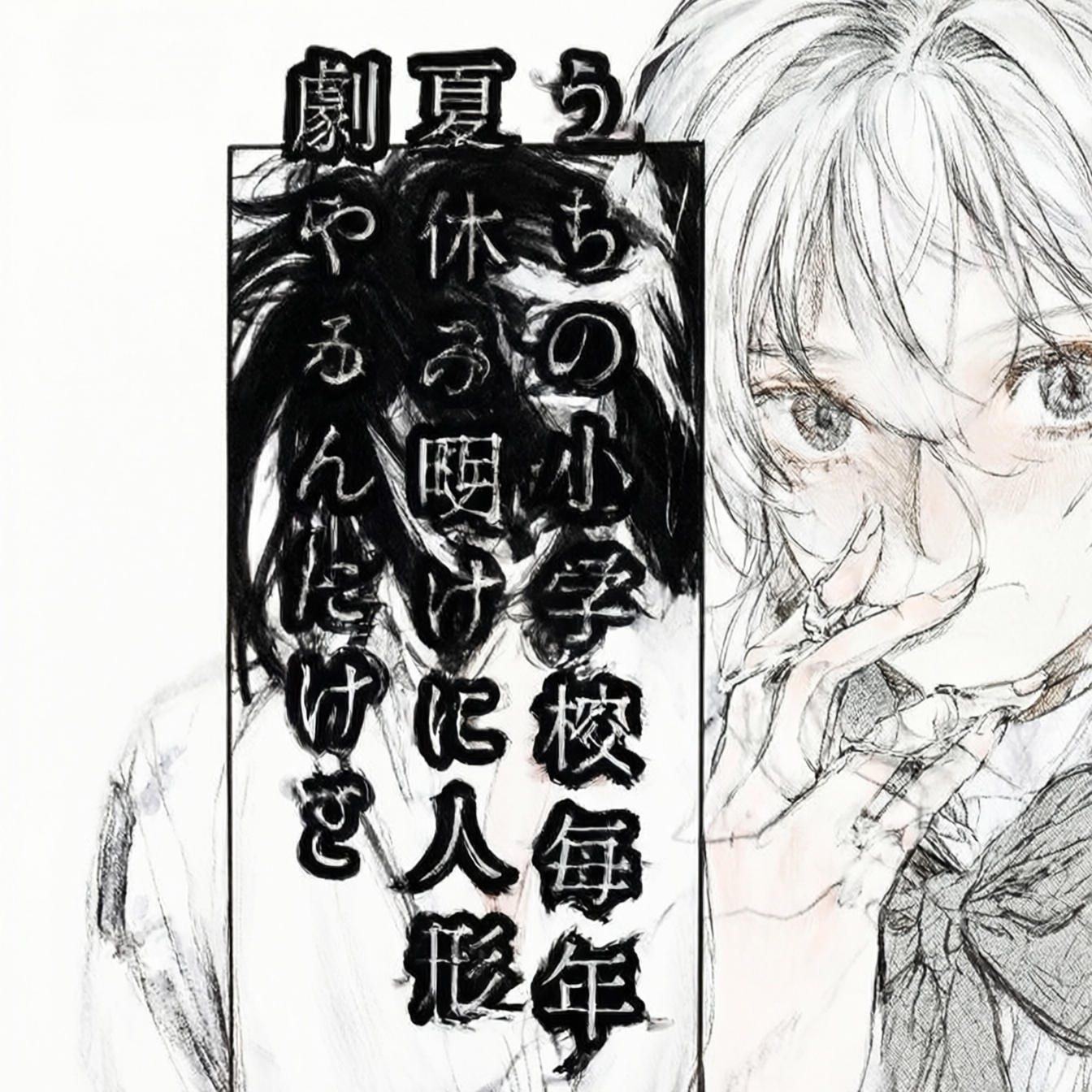}}\\[3pt]
        \small FlowChef~\cite{patel2025flowchef}
    \end{minipage}
    \hfill
    \begin{minipage}[t]{0.13\textwidth}
        \centering
        \fbox{\includegraphics[width=\linewidth]{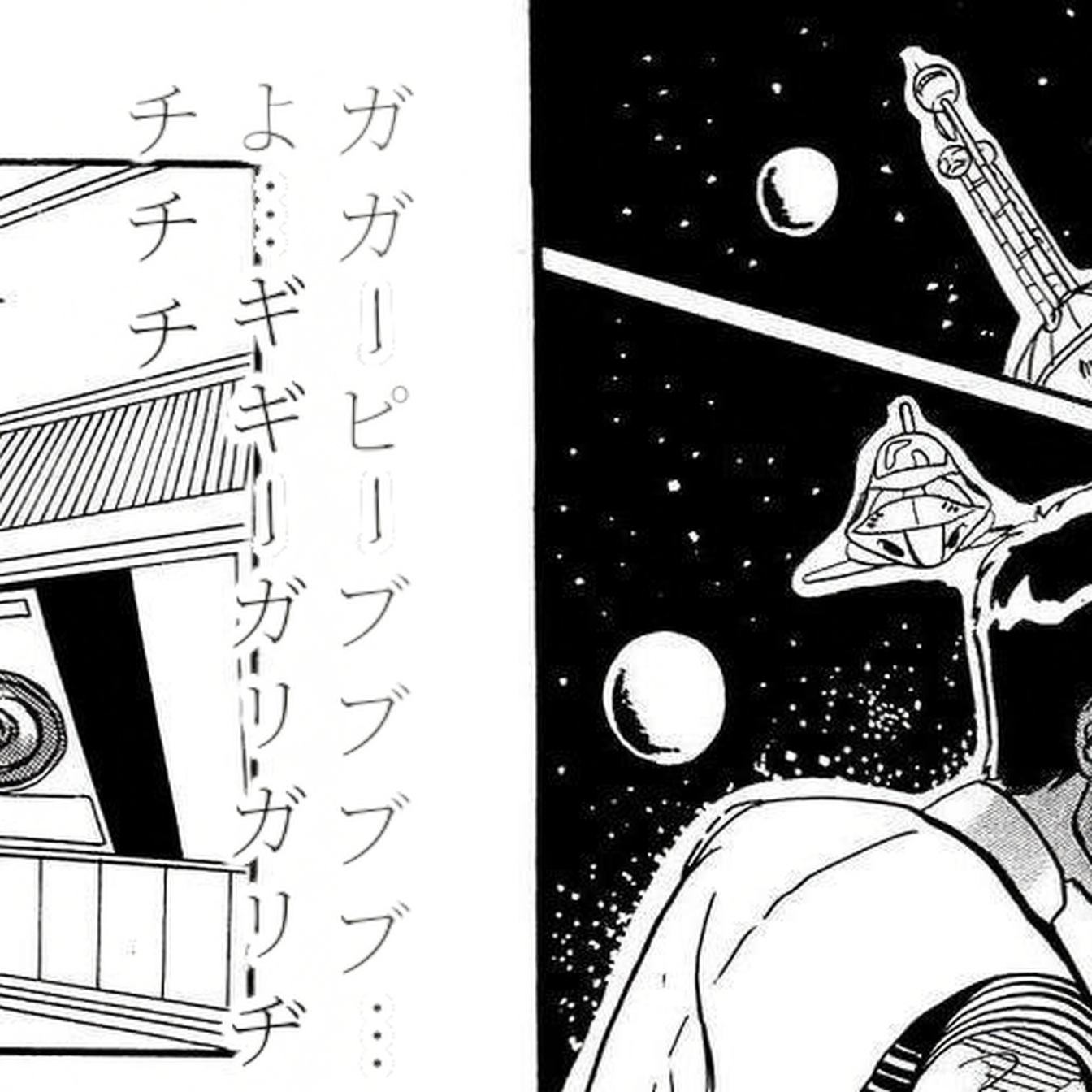}}\\[2pt]
        \fbox{\includegraphics[width=\linewidth]{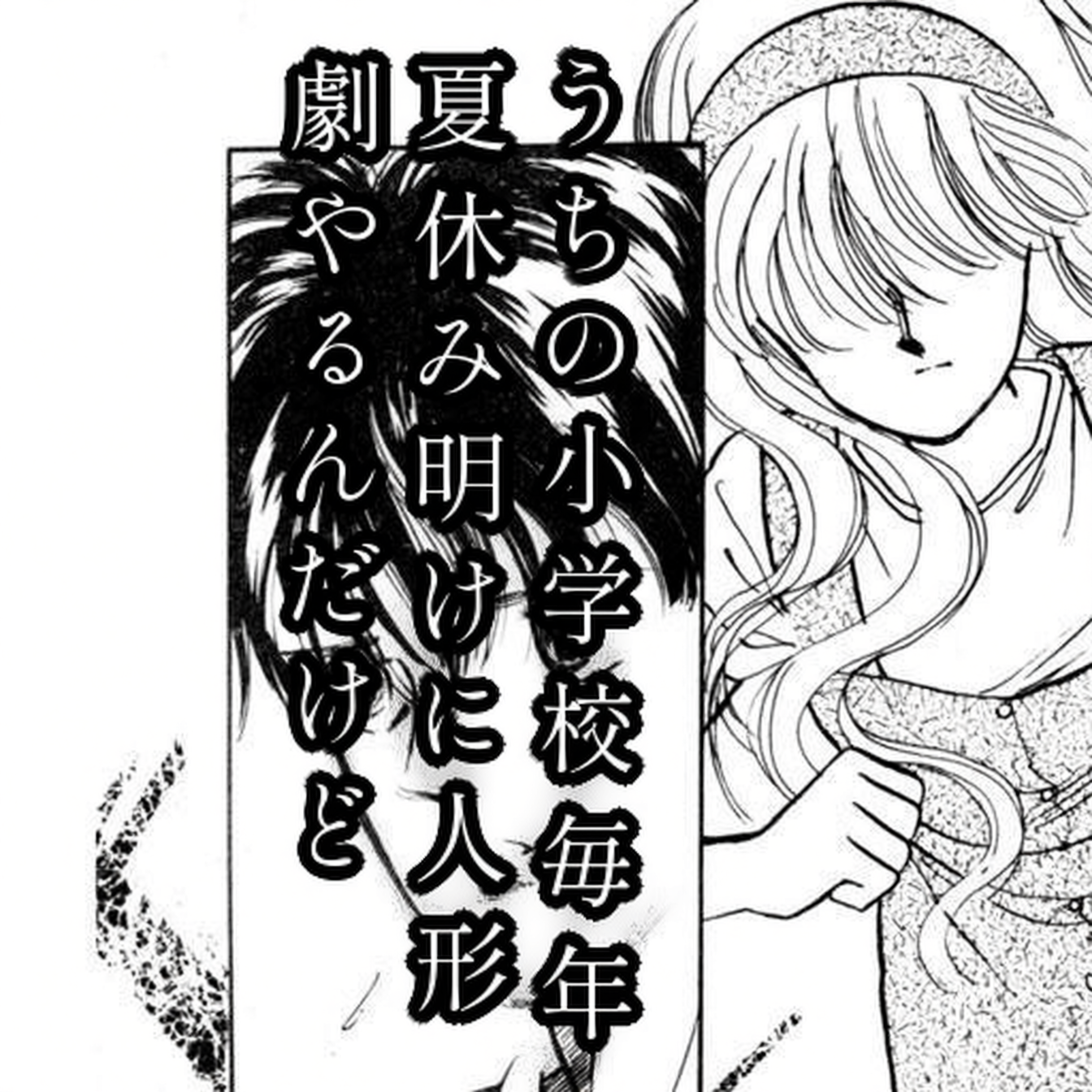}}\\[3pt]
        \small FlowEdit~\cite{kulikov2025flowedit}
    \end{minipage}
    \hfill
    \begin{minipage}[t]{0.13\textwidth}
        \centering
        \fbox{\includegraphics[width=\linewidth]{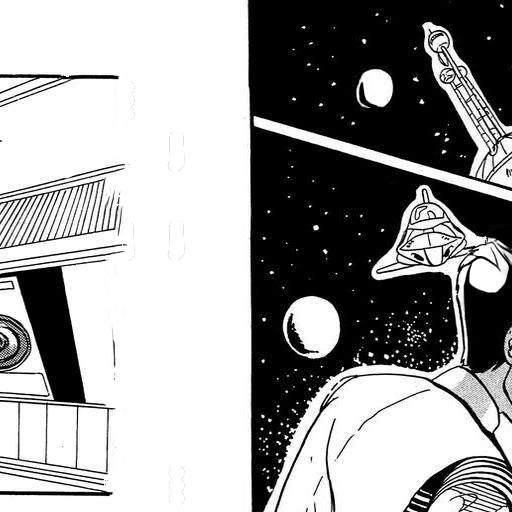}}\\[2pt]
        \fbox{\includegraphics[width=\linewidth]{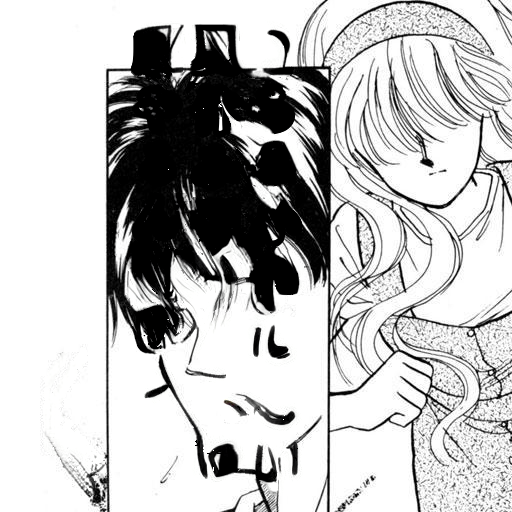}}\\[3pt]
        \small Sick-Zil~\cite{ko2020sickzil}
    \end{minipage}

    \caption{Qualitative comparison for text removal. \copyright Nagano Noriko \copyright Omi Ayuko}
    \label{fig:comparison_text_removal}
\end{figure*}

\subsection{Dataset}
Mask-free text removal aims to remove text written in an input manga image in a natural manner.
No segmentation mask for the text regions is provided at inference time.

Quantitative metrics such as PSNR, SSIM, and LPIPS require a text-free ground-truth image paired with each input.
However, no publicly available manga text-removal dataset provides such pairs while satisfying the copyright and reproducibility criteria detailed in supplementary Sec.~\ref{sec:evaluation_scope}.
We therefore construct controlled synthetic pairs from Manga109s.

Specifically, we constructed 3,204 pairs of input manga images with text and ground-truth manga images without text as follows.
First, we used the text annotations in Manga109s to extract the largest rectangular crop without annotated text regions from each two-page manga spread.
However, Manga109s also contains unannotated text such as handwritten text, so some cropped images could still include text.
We therefore manually inspected the crops and removed those that still contained text.
The resulting 3,204 cropped images were used as the ground-truth text-free manga images, and the corresponding input images with text were created by rendering text on top of them.
The rendered text content was randomly sampled from the text annotations originally present in the corresponding two-page spread before cropping.
The font was randomly selected from 24 fonts that looked natural for manga images.
Because the text is rendered synthetically, its exact pixel mask is known.
All text-region PSNR and SSIM values use this ground-truth rendering mask, rather than a mask estimated by SickZil-Machine or any other segmentation model; the evaluation is therefore independent of segmentation quality.

Although this construction does not fully reproduce real manga text, including handwritten effects, complex speech balloons, highly stylized fonts, and text printed over screentones, it provides exact paired ground truth and text-region masks.
Given the absence of suitable real paired data, we believe that this is a practical choice for fair and quantitative comparison.

\subsection{Implementation Details}
As the baseline, we used the open-source image editing model Qwen Image Edit 2509.
All hyperparameters were kept at their default values.
For our method, we set the number of corrected steps to $M=3$ and the correction strength to $\alpha=0.01$.
We use this fixed setting for all main experiments and do not tune either hyperparameter per image using ground truth.

\subsection{Preliminary Evaluation}
As a sanity check, we first verify that the proposed method faithfully reconstructs the input image when an empty prompt is given.
Specifically, we evaluate the discrepancy between the generated image under an empty prompt and {\bf the input image with rendered text}.
The results are shown in Table~\ref{tab:noprompt}.
Compared with the baseline, our method reconstructs the input image with text much more accurately.
The qualitative results in Fig.~\ref{fig:noprompt} further support this observation.

\subsection{Comparison with Other Methods}\label{sec:text_comparison}
We remove text by feeding an image with text and using the editing prompt ``Remove text. Keep everything else unchanged.'', and then measure the discrepancy between the generated edited image and {\bf the ground-truth text-free image}.
We do not include classification- or segmentation-oriented test-time adaptation methods as baselines: they optimize class predictions or segmentation outputs, whereas image editing requires generating an edited image and provides neither output type nor the corresponding adaptation objective, so they cannot be directly applied to this task.

We compare against the following methods:\\
{\bf Baseline}: directly applying Qwen Image Edit 2509.\\
{\bf Noise Inversion}: instead of generating the edited image from pure noise, this method starts generation from an intermediate point on the straight path connecting the noise $\bm{Z}_0$ and the input image $\bm{X}_{\mathrm{in}}$.
More specifically, generation starts from $(1-t_i)\bm{Z}_0 + t_i \bm{X}_{\mathrm{in}}$, which corresponds to advancing along the path for the first $i$ steps.
We report the best-performing setting, $i=1$.\\
{\bf FlowChef~\cite{patel2025flowchef} and FlowEdit~\cite{kulikov2025flowedit}}: image editing methods based on image generation models.
Their original implementations use the image generation model FLUX.1 [dev], but we also compare variants using Qwen Image 2512.
This makes the comparison better aligned with the image editing model Qwen Image Edit 2509 used by our method.
The source prompt is ``A manga image with text.'' and the target prompt is ``A manga image without text.''\\
{\bf SickZil-Machine~\cite{ko2020sickzil}}: a model designed and trained specifically for mask-free text removal in manga images.
We use the publicly available pretrained weights.\\
{\bf FireFlow~\cite{deng2025fireflow}, RF-Edit~\cite{wang2025taming}, and RF-Inversion~\cite{rout2025semantic}}: recent inversion and trajectory-based editing methods that we evaluate with FLUX.1 [dev] to provide a comparison with conceptually related trajectory methods.\\
Inpainting methods that require masks, such as Manga Inpainting~\cite{xie2021seamless}, cannot be directly applied to this task.

Table~\ref{tab:comparison_text_removal} reports the quantitative comparison.
For both Qwen Image Edit 2509 and FLUX.1 Kontext, our method obtains better values on all five metrics than the corresponding baseline.
This consistent improvement across the two image editors shows the benefit of the proposed correction over direct inference.
With FLUX.1 Kontext, our method further obtains the best values among all compared methods on four of the five metrics, including both text-region metrics.
With Qwen Image Edit 2509, our method also outperforms the task-specific SickZil-Machine within the exact ground-truth text regions, obtaining 20.04 versus 16.16 PSNR and 64.15 versus 48.8 SSIM.
These gaps are substantial in magnitude: the 3.88 dB PSNR difference corresponds to approximately $2.4\times$ lower MSE, and the SSIM difference is 15.35 points.
The text-region scores use the ground-truth rendering masks and are therefore independent of SickZil-Machine's estimated segmentation.
SickZil-Machine obtains higher full-image scores than the Qwen-based variant because it modifies only its segmented regions and copies all other pixels unchanged; consequently, the unedited background dominates the full-image metrics even when text is missed or local completion is inaccurate.
Results on additional image editing models are reported in supplementary Sec.~\ref{sec:model_applicability}.

Figure~\ref{fig:comparison_text_removal} shows qualitative comparisons of the edited images.
In the examples shown, the baseline edits regions more aggressively than necessary, whereas our method removes text while better preserving the overall structure of the input image.
In these examples, SickZil-Machine leaves text behind when text-region segmentation fails and does not faithfully complete some facial regions.

\subsection{Runtime and Memory}\label{sec:runtime}
We evaluate both runtime and peak memory using FLUX.1 Kontext on an NVIDIA RTX A6000 with 48 GB of VRAM.
We use FLUX.1 Kontext for this experiment because the baseline Qwen Image Edit 2509 pipeline, without the proposed correction, already exceeds the A6000's 48 GB memory capacity.
In the A6000/FLUX.1 Kontext configuration, runtime increases from 52.39 to 58.23 seconds, while peak memory changes from 33.83 to 33.85 GB, corresponding to 11\% runtime and 0.1\% peak-memory overhead.

As an additional latency measurement, we evaluate Qwen Image Edit 2509 on an NVIDIA H200.
The baseline takes an average of 35.56 seconds per image and our method takes 35.60 seconds over three images, with standard deviations of 0.01 and 0.03 seconds, respectively.
This corresponds to a 0.1\% runtime increase.

\subsection{Why Use an Empty Prompt}
\begin{figure}[t]
    \centering
    \setlength{\fboxrule}{0.5pt}
    \setlength{\fboxsep}{0pt}

    \begin{minipage}[b]{0.32\linewidth}
        \centering
        \fbox{\includegraphics[width=\linewidth]{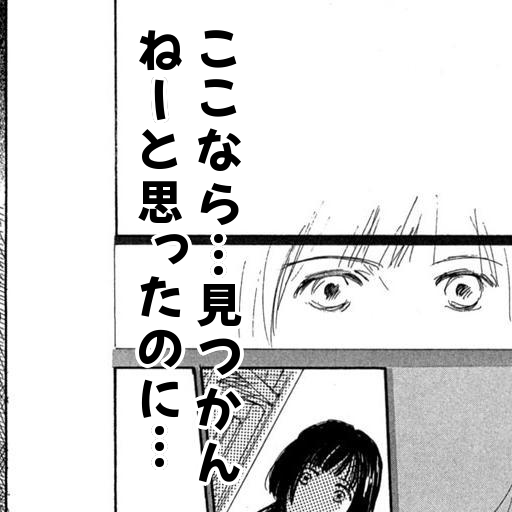}}\\[2pt]
        Input image
    \end{minipage}
    \hfill
    \begin{minipage}[b]{0.32\linewidth}
        \centering
        \fbox{\includegraphics[width=\linewidth]{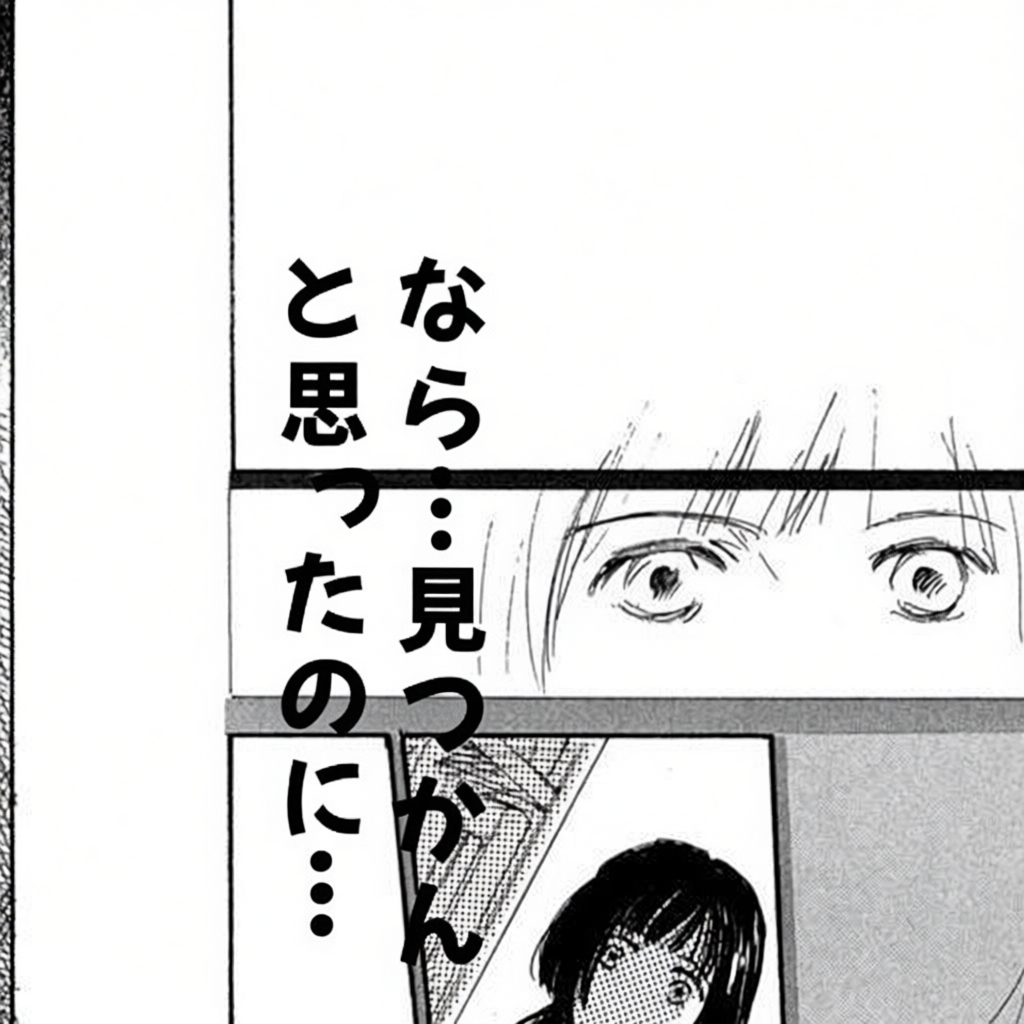}}\\[2pt]
        Edit prompt
    \end{minipage}
    \hfill
    \begin{minipage}[b]{0.32\linewidth}
        \centering
        \fbox{\includegraphics[width=\linewidth]{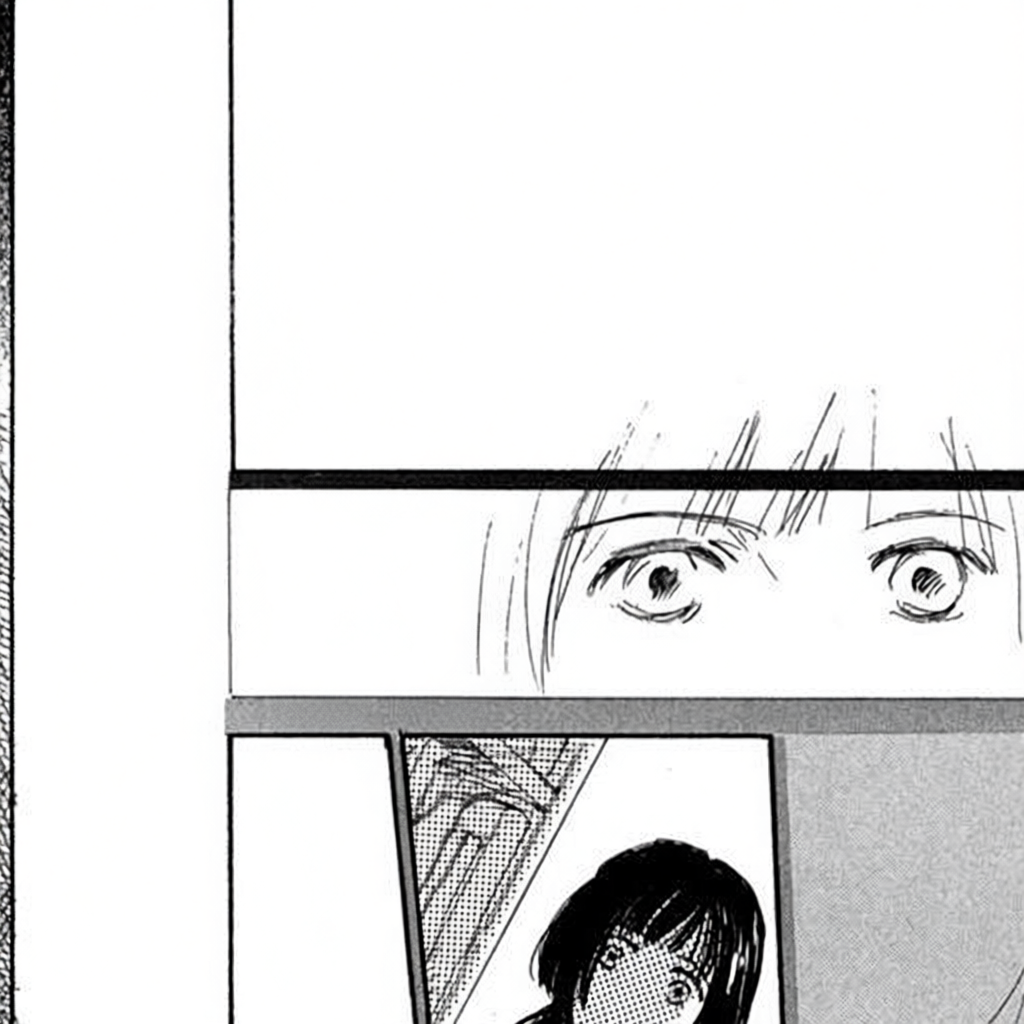}}\\[2pt]
        Empty prompt
    \end{minipage}
    \caption{Effect of the prompt used for trajectory correction. \copyright Okuda Momoko}
    \label{fig:effect}
\end{figure}

To verify the effect of using an empty prompt, we evaluate a variant of Eq.~(\ref{eq:opt_latent}) in which the edit prompt $c$ is used instead of the empty prompt $\emptyset$.
For all results in this comparison, we measure the discrepancy between the generated edited image and the ground-truth text-free image.
Table~\ref{tab:effect} shows the quantitative results.
When trajectory correction is performed using the edit prompt, the error inside the text region increases.
This happens because the generated result is corrected to be closer to the input image, causing the output to stay too close to the original image and thus leaving text behind (Fig.~\ref{fig:effect}).

\begin{table}[t]
  \centering
  \caption{Effect of the prompt used for trajectory correction}
  \tablebodyfont
  \setlength{\tabcolsep}{0.5mm}
  \begin{tabular}{@{}lrrrrr@{}}
    \toprule
    \multirow{2}{*}{Method} & \multicolumn{3}{c}{Full image} & \multicolumn{2}{c}{Text region} \\
    \cmidrule(lr){2-4} \cmidrule(lr){5-6}
     & PSNR $\uparrow$ & SSIM $\uparrow$ & LPIPS $\downarrow$ & PSNR $\uparrow$ & SSIM $\uparrow$ \\
    \midrule
    Baseline & 19.62 & 72.76 & 7.16 & 19.07 & 60.77 \\
    Edit prompt & 21.57 & 80.68 & 6.74 & 18.69 & 61.99 \\
    Empty prompt & \textbf{22.07} & \textbf{80.61} & \textbf{6.00} & \textbf{20.04} & \textbf{64.15} \\
    \bottomrule
  \end{tabular}
  \label{tab:effect}
\end{table}

\section{Conclusion}
We presented a lightweight, training-free trajectory correction that anchors early editing steps to empty-prompt reconstruction, preserving global composition during local manga edits.
Our exact one-step analysis shows how the correction moves the reconstructed endpoint toward the input without weakening the editing response introduced by the prompt.
Experiments with Qwen Image Edit 2509 and FLUX.1 Kontext indicate improved mask-free text removal over direct inference, with small practical overhead and no training, architectural changes, or additional sequential model passes.
Future work will examine real text-removal data, other applications such as colorization, and adaptive anchoring for edits requiring larger structural or semantic changes.

\FloatBarrier
\putbib[myref]
\end{bibunit}

\clearpage
\twocolumn[
\begin{center}
{\LARGE\bfseries Supplementary Material\par}
\vspace{0.5em}
{\Large\bfseries Inference-time Trajectory Optimization\\
for Structure-Preserving Manga Image Editing\par}
\vspace{1.5em}
\end{center}
]
\begin{bibunit}[aaai2027]
\appendix
\section{Evaluation Scope}\label{sec:evaluation_scope}
Our evaluation is designed around practical manga-editing tasks and data that satisfy both copyright-provenance and reproducibility criteria.
Because manga images are copyrighted artworks, we do not regard online availability alone as authorization to use commercial manga for training or evaluation.
Under our dataset-selection policy, we require documented permission for research use.
Manga109s~\cite{multimedia_aizawa_2020,mtap_matsui_2017} meets this criterion because its manga authors formally granted such permission, and its distribution supports reproducible evaluation.

We evaluate text removal and screentone synthesis using Manga109s.
These practically important tasks are complementary: text removal deletes existing content, whereas screentone synthesis adds new content to line art.
Both require the global manga structure to be preserved, and controlled preprocessing of Manga109s allows pairs of input and ground-truth images needed for quantitative evaluation to be constructed.

These criteria also guide our treatment of other datasets.
For existing large-scale paired colorization data~\cite{golyadkin2025closing}, the available documentation did not establish copyright provenance to the level required by our policy.
MangaZero~\cite{wu2025diffsensei} presents a separate reproducibility issue: its public release relies on source-image URLs, some of which are unavailable, making complete reconstruction of the released dataset difficult.
We therefore do not use either source for training or evaluation.
Semantic edits such as changing character attributes pose a different challenge because objective paired ground truth is difficult to construct.
The selected tasks thus provide reproducible quantitative benchmarks for both removing and adding content while preserving global manga structure.

\section{Additional Experiments: Mask-Free Text Removal}
Unless otherwise noted, all evaluations in this section measure the discrepancy between the generated edited image and the ground-truth text-free image.

\subsection{Effect of Optimized Trajectory Computation}
We next analyze the effect of the optimized trajectory computed by Eqs.~(\ref{eq:optimization}) and (\ref{eq:opt_latent}).
Instead of Eqs.~(\ref{eq:optimization}) and (\ref{eq:opt_latent}), we also consider correcting the trajectory using the straight path between the noise $\bm{Z}_0$ and the input image $\bm{X}_{\mathrm{in}}$:
\begin{equation}
    \bm{Z}^*_{t_i}=(1-t_i)\bm{Z}_0 + t_i \bm{X}_{\mathrm{in}}. \label{eq:straight_pass}
\end{equation}
Trajectory correction is then performed in the same way as Eq.~(\ref{eq:fixation}).

Table~\ref{tab:trajectory} summarizes the results.
Using the optimized trajectory yields better point estimates on all five metrics than correcting the trajectory with the straight path.
Because Rectified Flow models are trained using straight paths as supervision, the optimized trajectory should coincide with the straight path if a pretrained Rectified Flow model behaves ideally.
However, on data that are rarely seen during training, such as manga images, Rectified Flow models may deviate from this ideal behavior, making the optimized trajectory more effective in the evaluated setting.

\begin{table}[t]
  \centering
  \caption{Effect of the optimized trajectory}
  \tablebodyfont
  \setlength{\tabcolsep}{0.5mm}
  \begin{tabular}{@{}lrrrrr@{}}
    \toprule
    \multirow{2}{*}{Method} & \multicolumn{3}{c}{Full image} & \multicolumn{2}{c}{Text region} \\
    \cmidrule(lr){2-4} \cmidrule(lr){5-6}
     & PSNR $\uparrow$ & SSIM $\uparrow$ & LPIPS $\downarrow$ & PSNR $\uparrow$ & SSIM $\uparrow$ \\
    \midrule
    Baseline & 19.62 & 72.76 & 7.16 & 19.07 & 60.77 \\
    Straight path & 19.69 & 73.13 & 7.74 & 19.13 & 61.12 \\
    Optimized path & \textbf{22.07} & \textbf{80.61} & \textbf{6.00} & \textbf{20.04} & \textbf{64.15} \\
    \bottomrule
  \end{tabular}
  \label{tab:trajectory}
\end{table}

\subsection{Effect of Hyperparameters}
Table~\ref{tab:hypara_text_removal} serves two purposes: it verifies stability within a practical operating range and probes deliberately extreme settings to clarify the role of each hyperparameter.
Within the practical ranges examined, the proposed method improves all five metrics over the baseline for all tested values of $M$ between 1 and 7 and of $\alpha$ between 0.01 and 0.05.
The method therefore does not require sensitive tuning within these ranges; stable improvements are obtained by correcting a small number of early steps and keeping $\alpha$ within the same order of magnitude as the default.

The settings below the separator are deliberately outside the practical ranges and are included to interpret the role of each hyperparameter, rather than as recommended operating points.
Setting $M=9$ extends the correction into later timesteps and makes the output overly close to the input: full-image PSNR and SSIM improve, but residual text degrades the text-region metrics.
This result illustrates why the method is designed to correct only the early trajectory.
Similarly, $\alpha=0.001$ and $\alpha=0.1$ are respectively one tenth and ten times the default value.
The former is too weak to sufficiently correct the trajectory, whereas the latter over-corrects toward the input and degrades LPIPS and the text-region metrics.
These deliberately extreme values illustrate the expected under- and over-correction behavior; they do not indicate fragility within the practical range.

For all main experiments, we use the same fixed setting, $M=3$ and $\alpha=0.01$, without per-image tuning using ground truth.

\begin{table}[t]
  \centering
  \caption{Stability within practical ranges and interpretation using deliberately extreme hyperparameter settings}
  \tablebodyfont
  \setlength{\tabcolsep}{0.5mm}
  \begin{tabular}{@{}llrrrrr@{}}
    \toprule
    \multirow{2}{*}{Method} & Hyper- & \multicolumn{3}{c}{Full image} & \multicolumn{2}{c}{Text region} \\
    \cmidrule(lr){3-5} \cmidrule(lr){6-7}
     & param. & PSNR $\uparrow$ & SSIM $\uparrow$ & LPIPS $\downarrow$ & PSNR $\uparrow$ & SSIM $\uparrow$ \\
    \midrule
    Baseline & & 19.62 & 72.76 & 7.16 & 19.07 & 60.77 \\
    \midrule
    \multicolumn{7}{@{}l}{\textit{Practical operating range}} \\
    \midrule
    \multirow{9}{*}{Ours} & $M=1$ & 20.89 & 76.87 & 6.94 & 19.71 & 62.56 \\ \cmidrule(lr){2-7}
    & $M=3$ & 22.07 & 80.61 & 6.00 & 20.04 & 64.15 \\ \cmidrule(lr){2-7}
    & $M=5$ & 22.31 & 81.67 & 6.01 & 19.82 & 63.59 \\ \cmidrule(lr){2-7}
    & $M=7$ & 22.37 & 82.42 & 6.45 & 19.23 & 61.31 \\ \cmidrule(lr){2-7}
    & $\alpha=0.01$ & 22.07 & 80.61 & 6.00 & 20.04 & 64.15 \\ \cmidrule(lr){2-7}
    & $\alpha=0.02$ & 23.40 & 84.38 & 5.30 & 20.54 & 64.97 \\ \cmidrule(lr){2-7}
    & $\alpha=0.03$ & 24.39 & 87.43 & 5.10 & 20.76 & 64.49 \\ \cmidrule(lr){2-7}
    & $\alpha=0.04$ & 24.73 & 89.03 & 5.30 & 20.36 & 62.60 \\ \cmidrule(lr){2-7}
    & $\alpha=0.05$ & 24.87 & 89.84 & 5.46 & 20.25 & 61.65 \\
    \midrule
    \multicolumn{7}{@{}l}{\textit{Deliberately extreme settings}} \\
    \midrule
    \multirow{3}{*}{Ours} & $M=9$ & 22.38 & 83.10 & 6.93 & 18.48 & 57.92 \\ \cmidrule(lr){2-7}
    & $\alpha=0.001$ & 19.67 & 72.92 & 7.87 & 19.08 & 60.83 \\ \cmidrule(lr){2-7}
    & $\alpha=0.1$ & 22.74 & 86.64 & 9.76 & 18.40 & 54.58 \\
    \bottomrule
  \end{tabular}
  \label{tab:hypara_text_removal}
\end{table}

\subsection{Applicability to Various Image Editing Models}\label{sec:model_applicability}
Table~\ref{tab:models_text_removal} compares the correction across all four image editing backbones evaluated for text removal.
The correction improves all five metrics for Qwen Image Edit 2509 and FLUX.1 Kontext.
For Qwen Image Edit 2511 and FLUX.2 Klein 9B, four of the five metrics improve only slightly, while text-region SSIM and PSNR, respectively, decrease slightly.
These results show that the correction can be applied to multiple backbones without model-specific training, although the magnitude of the quantitative changes varies across them.

\begin{table*}[t]
  \centering
  \caption{Evaluation across image editing models for text removal}
  \tablebodyfont
  \setlength{\tabcolsep}{0.5mm}
  \begin{tabular}{@{}llrrrrr@{}}
    \toprule
    \multirow{2}{*}{Method} & \multirow{2}{*}{Model} & \multicolumn{3}{c}{Full image} & \multicolumn{2}{c}{Text region} \\
    \cmidrule(lr){3-5} \cmidrule(lr){6-7}
     & & PSNR $\uparrow$ & SSIM $\uparrow$ & LPIPS $\downarrow$ & PSNR $\uparrow$ & SSIM $\uparrow$ \\
    \midrule
    Baseline & Qwen Image Edit 2509 & 19.62 & 72.76 & 7.16 & 19.07 & 60.77 \\
    Ours & Qwen Image Edit 2509 & \textbf{22.07} & \textbf{80.61} & \textbf{6.00} & \textbf{20.04} & \textbf{64.15} \\ \midrule
    Baseline & Qwen Image Edit 2511 & 21.19 & 82.63 & 10.16 & 18.51 & \textbf{62.86} \\
    Ours & Qwen Image Edit 2511 & \textbf{21.39} & \textbf{83.05} & \textbf{9.84} & \textbf{18.54} & 62.55 \\ \midrule
    Baseline & FLUX.1 Kontext [dev] & 26.42 & 90.47 & 6.60 & 21.00 & 66.97 \\
    Ours & FLUX.1 Kontext [dev] & \textbf{27.36} & \textbf{93.81} & \textbf{4.21} & \textbf{21.48} & \textbf{68.97} \\ \midrule
    Baseline & FLUX.2 Klein 9B & 26.90 & 94.67 & 3.48 & \textbf{19.72} & 66.32 \\
    Ours & FLUX.2 Klein 9B & \textbf{27.07} & \textbf{94.91} & \textbf{3.44} & 19.61 & \textbf{66.38} \\
    \bottomrule
  \end{tabular}
  \label{tab:models_text_removal}
\end{table*}

\section{Experiments: Screentone Synthesis}
\subsection{Dataset}\label{sec:screentone_data}
Screentone synthesis takes a line-art image as input and outputs a final manga image with screentones.
Following \citet{tsubota2019synthesis}, we constructed 1,355 pairs of input line art and ground-truth manga images with screentones as follows.
We cropped character bounding boxes from six titles in Manga109s to obtain 1,355 ground-truth manga images with screentones.
We then converted them into input line-art images using Manga Line Extraction~\cite{li2017deep}.

\subsection{Comparison Methods}
As the baseline model, we use Qwen Image Edit 2511.
The editing prompt is ``Convert the line art into a finished Japanese black and white monochrome manga image.''

We additionally compare with Noise Inversion, FlowChef~\cite{patel2025flowchef}, and FlowEdit~\cite{kulikov2025flowedit}.
For FlowChef and FlowEdit, the source prompt is ``A line art image.'' and the target prompt is ``A finished Japanese black and white monochrome manga image.''

The task-specific methods most directly relevant to this setting are Sketch2Manga~\cite{lin2024sketch2manga} and the earlier method of \citet{tsubota2019synthesis}.
Sketch2Manga is the more recent method and provides public pretrained weights, enabling reproducible evaluation.
The method of Tsubota et al. requires screentone pattern data at inference time, but these data remained unavailable even after we contacted the authors.
We therefore use Sketch2Manga as the task-specific baseline; as the more recent and reproducible of the two methods, it provides an appropriate comparison with a model designed specifically for screentone synthesis.

\subsection{Results}
The proposed correction is not designed to improve the base editor's screentone-synthesis ability itself.
Its role is to suppress unintended structural changes, thereby preserving the input's global composition, including its line-art structure, while the base editor performs screentone synthesis.

Quantitatively evaluating this effect is challenging.
Screentone synthesis has no unique correct output because many screentone placements can produce a plausible finished manga image, and the ground-truth image constructed in Sec.~\ref{sec:screentone_data} is only one such realization.
Paired pixel-wise metrics such as PSNR and SSIM are therefore unsuitable.
Successful outputs should nevertheless resemble the distribution of finished ground-truth manga.
We therefore adopt FID and CMMD~\cite{jayasumana2024rethinking}, which are commonly used to evaluate image generation, as distribution-level proxies for this task.
However, these metrics do not measure whether each output preserves the structure of its input; because the correction does not target screentone-synthesis quality itself, its intended effect may not produce a large change in FID or CMMD.
Since no standard metric simultaneously evaluates input-structure preservation and screentone quality, we use qualitative comparisons to directly assess both aspects.

Table~\ref{tab:comparison_screentone_removal} shows the quantitative comparison.
For Qwen Image Edit 2511, the changes from 36.08 to 36.07 FID and from 0.3140 to 0.3129 CMMD are marginal, and Noise Inversion obtains a comparable, slightly better CMMD of 0.3123.
In contrast, the proposed correction reduces both FID and CMMD for FLUX.1 Kontext and FLUX.2 Klein 2B, with the largest change being a 9.28-point FID reduction for FLUX.1 Kontext.
One possible explanation is that the FLUX-family baselines have higher FID and CMMD than Qwen Image Edit 2511, leaving more room for improvement; in these settings, suppressing structural deviations may also move the outputs closer to the ground-truth distribution.
Because FID and CMMD do not isolate structure preservation, we treat this as a plausible interpretation rather than a definitive causal conclusion.

\begin{table}[t]
  \centering
  \caption{Quantitative evaluation of screentone synthesis}
  \tablebodyfont
  \setlength{\tabcolsep}{0.5mm}
  \begin{tabular}{@{}llrr@{}}
    \toprule
    Method & Model & FID $\downarrow$ & CMMD $\downarrow$ \\
    \midrule
    Baseline & Qwen Image Edit 2511 & 36.08 & 0.3140 \\
    Noise Inversion & Qwen Image Edit 2511 & 36.71 & \textbf{0.3123} \\
    Ours & Qwen Image Edit 2511 & \textbf{36.07} & 0.3129 \\
    FlowChef & Qwen Image 2512 & 81.88 & 0.9993 \\
    FlowEdit & Qwen Image 2512 & 61.32 & 0.4669 \\ \midrule
    Baseline & FLUX.1 Kontext [dev] & 92.97 & 0.3746 \\
    Ours & FLUX.1 Kontext [dev] & 83.69 & 0.3704 \\
    FlowChef & FLUX.1 [dev] & 93.46 & 0.3928 \\
    FlowEdit & FLUX.1 [dev] & 68.28 & 1.3403 \\ \midrule
    Baseline & FLUX.2 Klein 2B & 47.22 & 0.5630 \\
    Ours & FLUX.2 Klein 2B & 46.89 & 0.5547 \\ \midrule
    Sketch2Manga & Finetuned SD1.5 & 139.09 & 2.1467 \\
    \bottomrule
  \end{tabular}
  \label{tab:comparison_screentone_removal}
\end{table}

\begin{figure*}[t]
    \centering
    \setlength{\fboxrule}{0.5pt}
    \setlength{\fboxsep}{0pt}

    \begin{minipage}[t]{0.13\textwidth}
        \centering
        \fbox{\includegraphics[width=\linewidth]{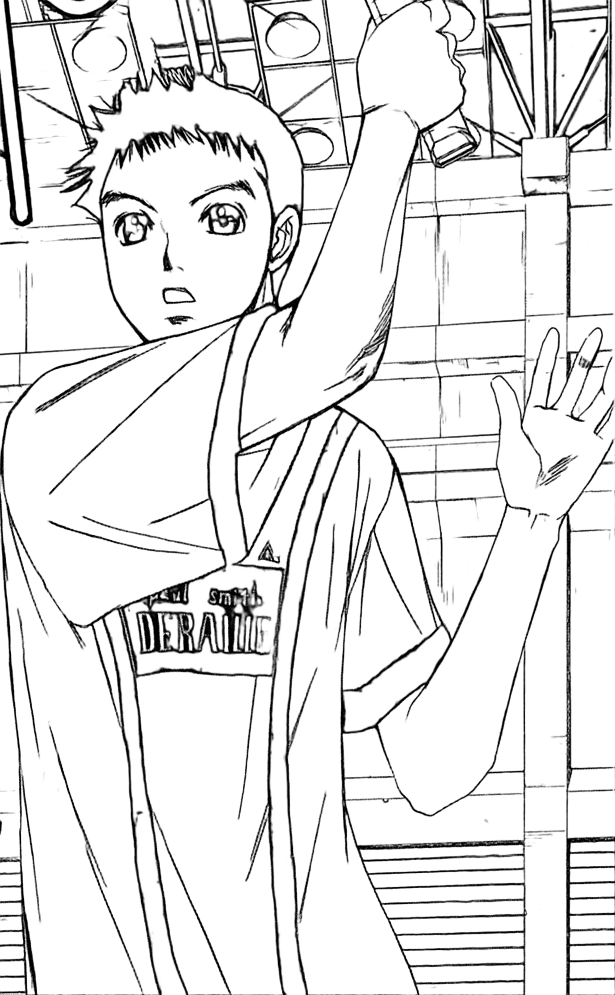}}\\[2pt]
        \fbox{\includegraphics[width=\linewidth]{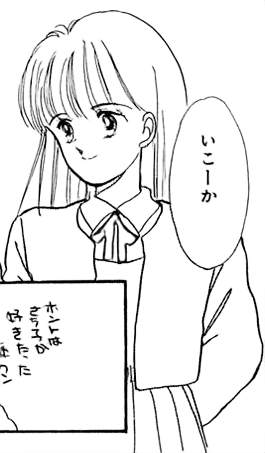}}\\[3pt]
        \small Input
    \end{minipage}
    \hfill
    \begin{minipage}[t]{0.13\textwidth}
        \centering
        \fbox{\includegraphics[width=\linewidth]{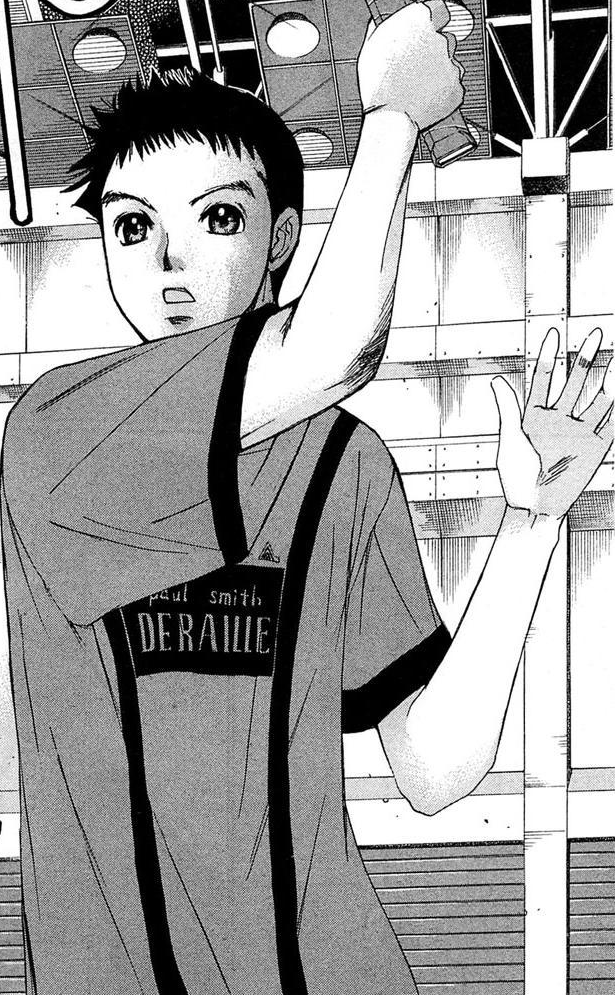}}\\[2pt]
        \fbox{\includegraphics[width=\linewidth]{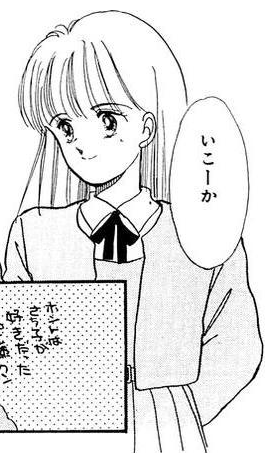}}\\[3pt]
        \small GT
    \end{minipage}
    \hfill
    \begin{minipage}[t]{0.13\textwidth}
        \centering
        \fbox{\includegraphics[width=\linewidth]{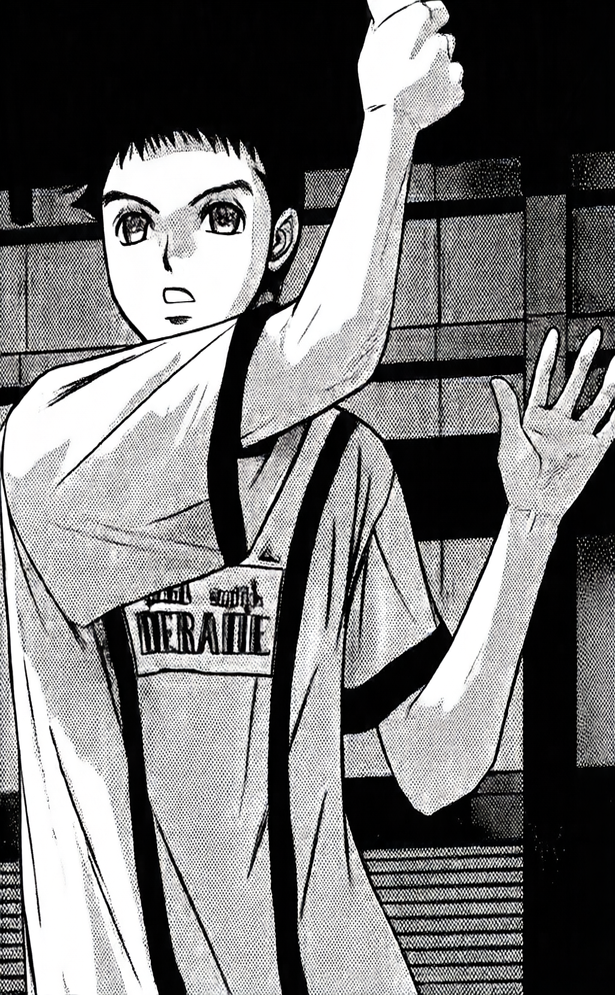}}\\[2pt]
        \fbox{\includegraphics[width=\linewidth]{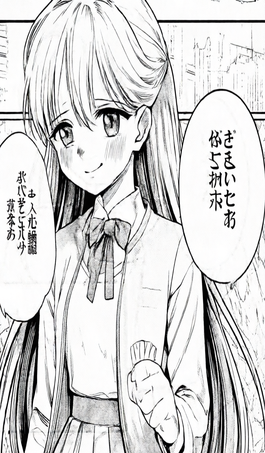}}\\[3pt]
        \small Baseline
    \end{minipage}
    \hfill
    \begin{minipage}[t]{0.13\textwidth}
        \centering
        \fbox{\includegraphics[width=\linewidth]{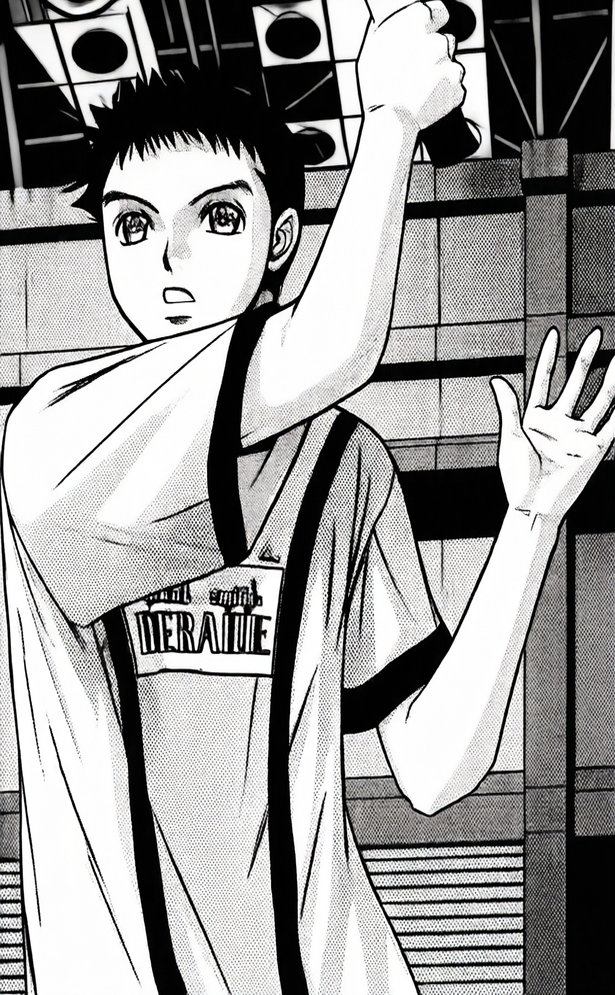}}\\[2pt]
        \fbox{\includegraphics[width=\linewidth]{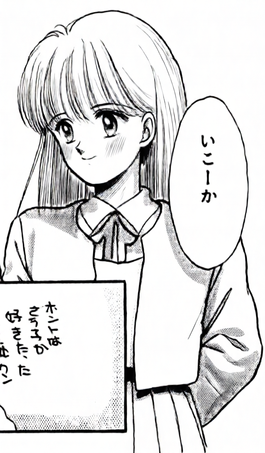}}\\[3pt]
        \small Ours
    \end{minipage}
    \hfill
    \begin{minipage}[t]{0.13\textwidth}
        \centering
        \fbox{\includegraphics[width=\linewidth]{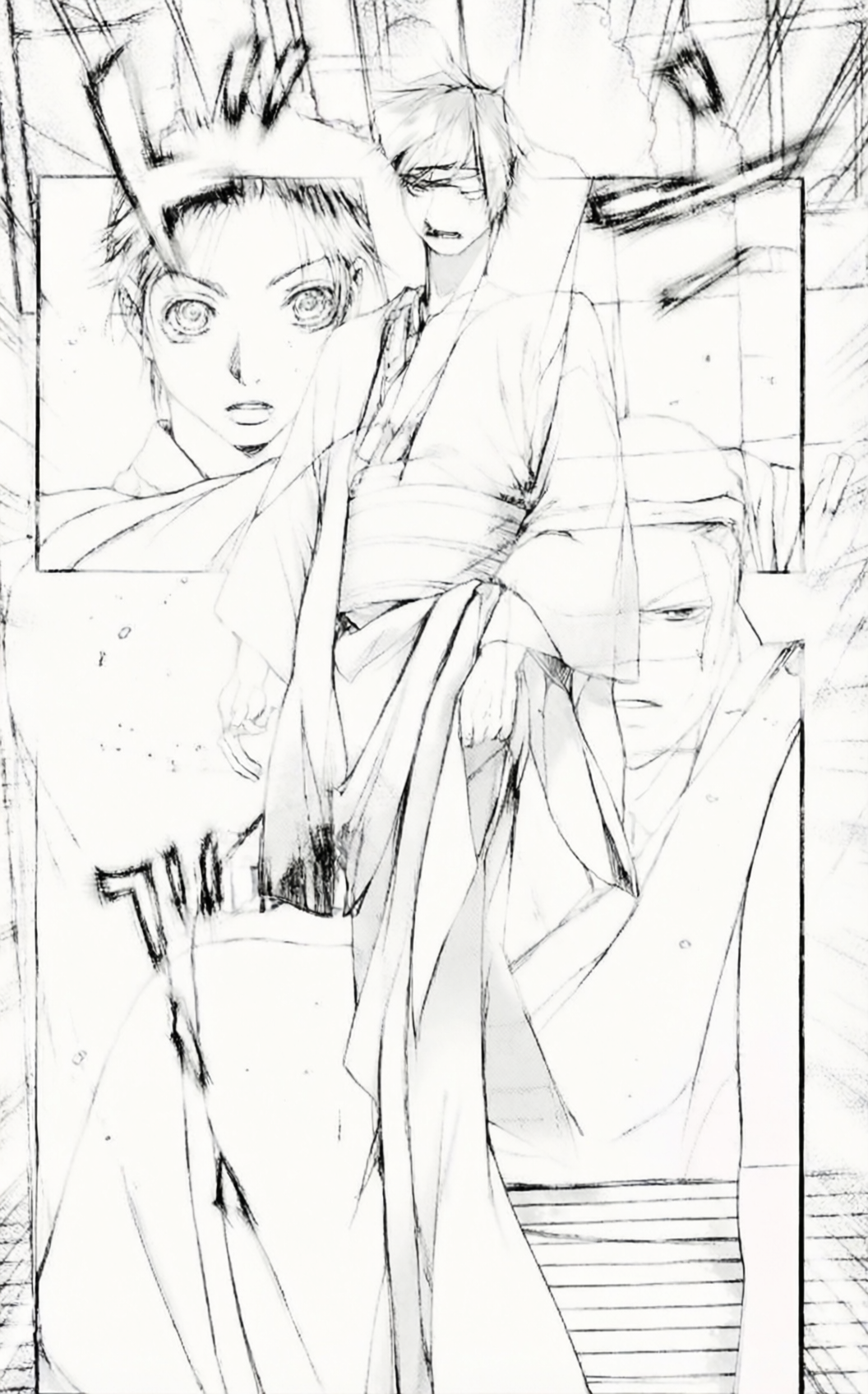}}\\[2pt]
        \fbox{\includegraphics[width=\linewidth]{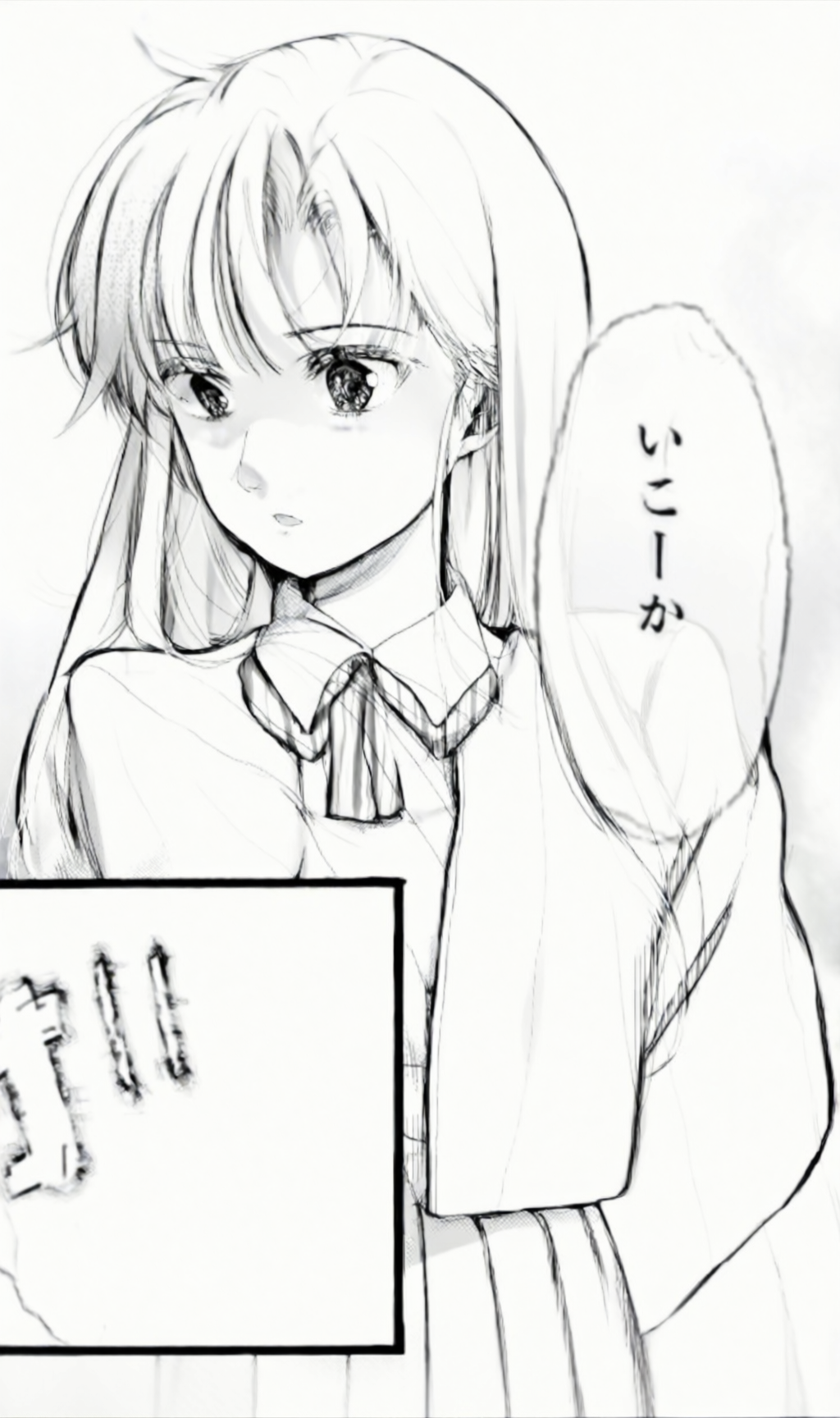}}\\[3pt]
        \small FlowChef~\cite{patel2025flowchef}
    \end{minipage}
    \hfill
    \begin{minipage}[t]{0.13\textwidth}
        \centering
        \fbox{\includegraphics[width=\linewidth]{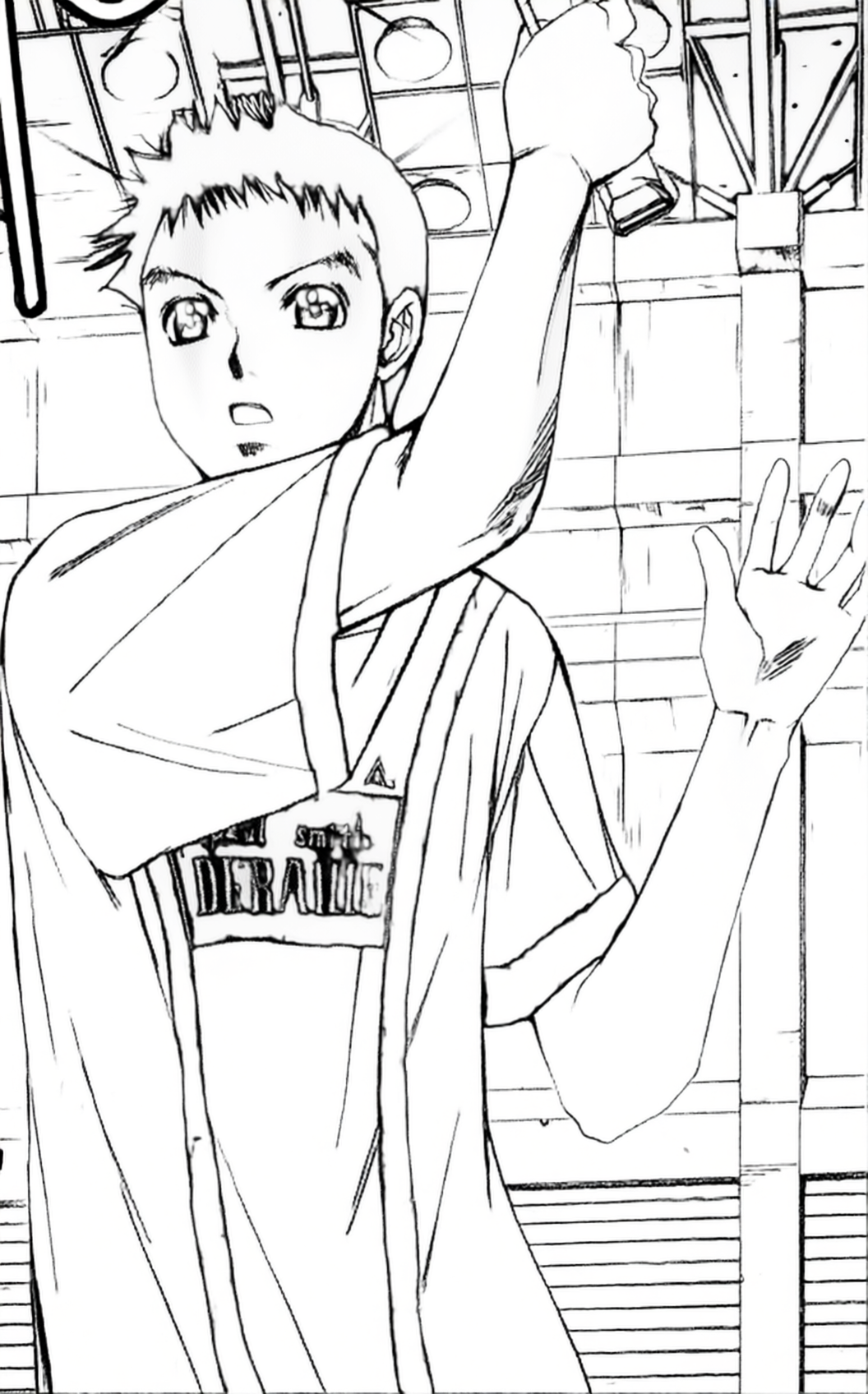}}\\[2pt]
        \fbox{\includegraphics[width=\linewidth]{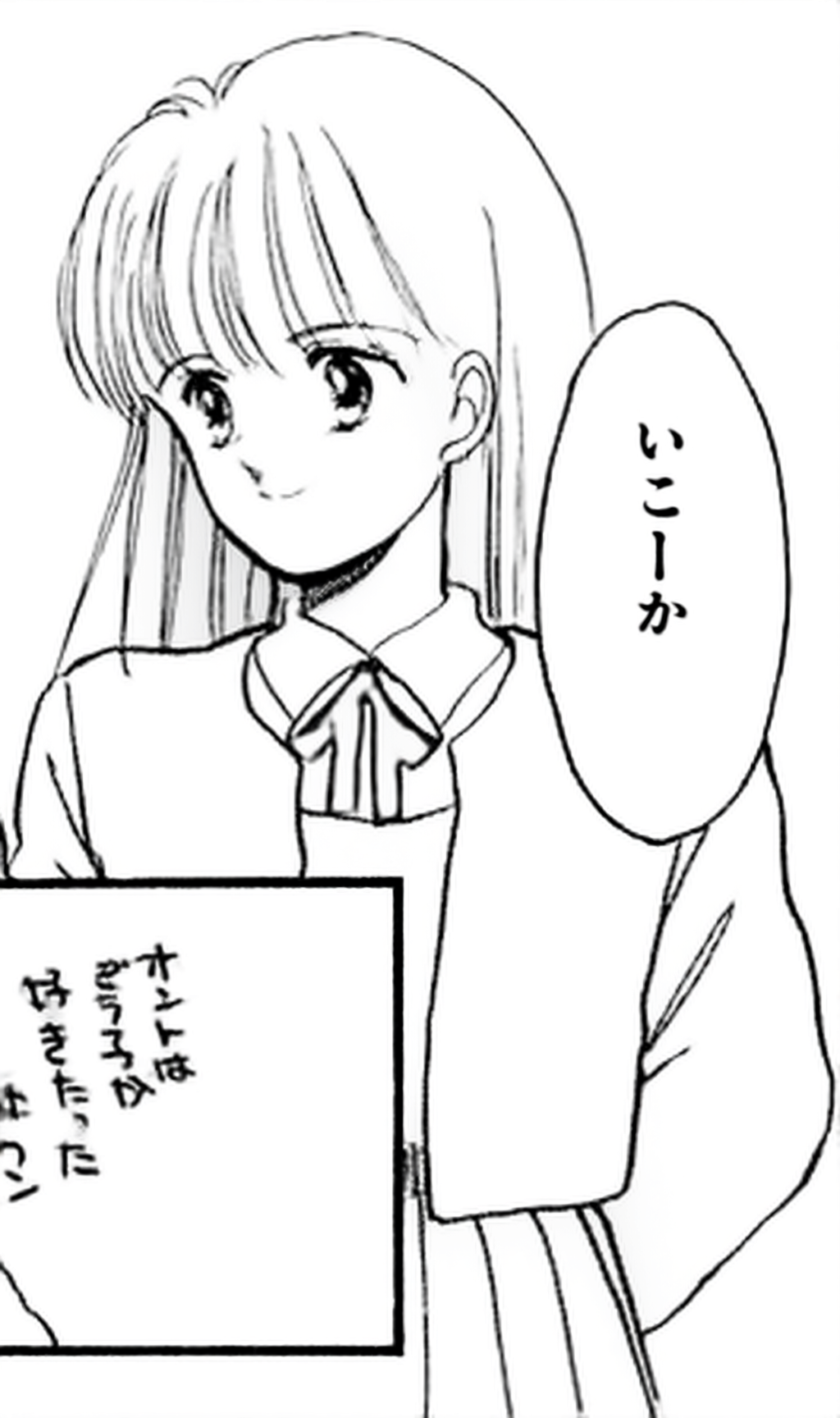}}\\[3pt]
        \small FlowEdit~\cite{kulikov2025flowedit}
    \end{minipage}
    \hfill
    \begin{minipage}[t]{0.13\textwidth}
        \centering
        \fbox{\includegraphics[width=\linewidth]{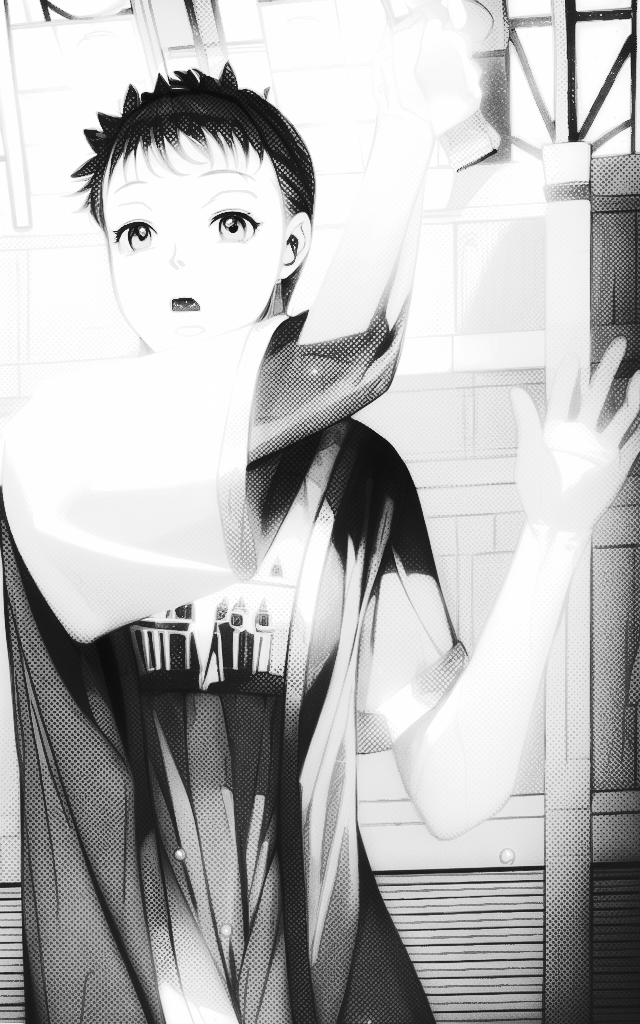}}\\[2pt]
        \fbox{\includegraphics[width=\linewidth]{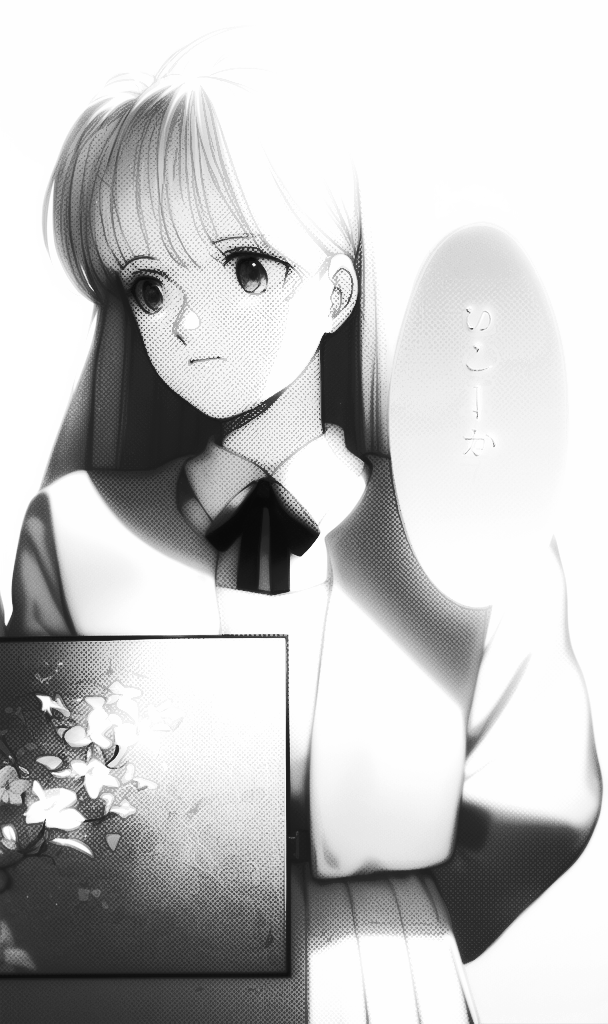}}\\[3pt]
        \small Sketch2Manga~\cite{lin2024sketch2manga}
    \end{minipage}

    \caption{Qualitative comparison for screentone synthesis. \copyright Saki Kaori \copyright Kurita Riku}
    \label{fig:comparison_screentone_generation}
\end{figure*}

Figure~\ref{fig:comparison_screentone_generation} provides qualitative comparisons.
In the examples shown, the proposed correction exhibits its intended behavior: it better preserves the input's global composition, including its line-art structure, while the base editor continues to produce natural screentones.
For Qwen Image Edit 2511, this qualitative structure-preservation effect is not reflected by the nearly unchanged FID and CMMD, illustrating why the distribution-level metrics alone are insufficient for evaluating the intended effect.

\FloatBarrier
\putbib[myref]
\end{bibunit}

\end{document}